\newif\ifisarxiv
\isarxivtrue
\newif\ifisneurips
\isneuripsfalse

\documentclass{article}

\usepackage{hyperref}
\usepackage{xcolor}
\hypersetup{
    colorlinks,
    linkcolor={red!40!gray},
    citecolor={blue!40!gray},
    urlcolor={blue!70!gray}
}
\usepackage{mdframed}

\usepackage[nonatbib,preprint]{neurips_2020_arxiv}
\usepackage[numbers]{natbib}

\usepackage[utf8]{inputenc}
\usepackage[T1]{fontenc}
\usepackage{url}
\usepackage{booktabs}
\usepackage{amsfonts}
\usepackage{nicefrac}
\usepackage{microtype}
\usepackage{amsmath,amssymb,amsthm}
\usepackage{mathtools}
\usepackage{color}
\usepackage{graphicx}
\usepackage{multirow}
\usepackage{wrapfig}
\usepackage{xfrac}
\usepackage{caption}
\usepackage{enumitem}
\usepackage{todonotes}
\usepackage{xspace}
\usepackage{newfloat}

\usepackage{algorithm}
\usepackage{algorithmicx}
\usepackage{algcompatible}
\algnewcommand\algorithmicreturn{\textbf{return} }
\algnewcommand\RETURN{\State \algorithmicreturn}
\usepackage{etoolbox}
\usepackage{tikz}
\usetikzlibrary{tikzmark}
\usetikzlibrary{calc}

\errorcontextlines\maxdimen

\newcommand{\ALGtikzmarkcolor}{black}
\newcommand{\ALGtikzmarkextraindent}{4pt}
\newcommand{\ALGtikzmarkverticaloffsetstart}{-.5ex}
\newcommand{\ALGtikzmarkverticaloffsetend}{-.5ex}
\makeatletter
\newcounter{ALG@tikzmark@tempcnta}

\newcommand\ALG@tikzmark@start{%
    \global\let\ALG@tikzmark@last\ALG@tikzmark@starttext%
    \expandafter\edef\csname ALG@tikzmark@\theALG@nested\endcsname{\theALG@tikzmark@tempcnta}%
    \tikzmark{ALG@tikzmark@start@\csname ALG@tikzmark@\theALG@nested\endcsname}%
    \addtocounter{ALG@tikzmark@tempcnta}{1}%
}

\def\ALG@tikzmark@starttext{start}
\newcommand\ALG@tikzmark@end{%
    \ifx\ALG@tikzmark@last\ALG@tikzmark@starttext
    \else
        \tikzmark{ALG@tikzmark@end@\csname ALG@tikzmark@\theALG@nested\endcsname}%
        \tikz[overlay,remember picture] \draw[\ALGtikzmarkcolor] let \p{S}=($(pic cs:ALG@tikzmark@start@\csname ALG@tikzmark@\theALG@nested\endcsname)+(\ALGtikzmarkextraindent,\ALGtikzmarkverticaloffsetstart)$), \p{E}=($(pic cs:ALG@tikzmark@end@\csname ALG@tikzmark@\theALG@nested\endcsname)+(\ALGtikzmarkextraindent,\ALGtikzmarkverticaloffsetend)$) in (\x{S},\y{S})--(\x{S},\y{E});%
    \fi
    \gdef\ALG@tikzmark@last{end}%
}

\apptocmd{\ALG@beginblock}{\ALG@tikzmark@start}{}{\errmessage{failed to patch}}
\pretocmd{\ALG@endblock}{\ALG@tikzmark@end}{}{\errmessage{failed to patch}}
\makeatother

\usepackage{setspace}

\DeclareFloatingEnvironment[
  fileext = loalg ,
  listname = {List of algorithms} ,
  name = AlgList
]{loalg}

\usepackage{cleveref}
\Crefformat{section}{Section~#2#1#3}
\Crefformat{proposition}{Proposition~#2#1#3}
\Crefformat{equation}{Equation~#2#1#3}
\Crefname{ALC@unique}{Line}{Lines}

\def\Vol{{\mathrm{R\textnormal{-}DPP}}}

\def\DPP{{\mathrm{DPP}}}
\def\PB{{\mathrm{PB}}}

\def\poly{{\mathrm{poly}}}

\def\simiid{\overset{\textnormal{\fontsize{6}{6}\selectfont
i.i.d.}}{\sim}}

\newcommand{\transp}{\top}

\newcommand{\normsmall}[1]{\Vert #1 \Vert}

\newcommand{\CommaBin}{\mathbin{\raisebox{0.5ex}{,}}}

\def\sigmat{\widetilde{\sigma}}

\newcommand{\cO}{\mathcal{O}}

\def\W{\mathbf W}

\def\Lb{\mathbf{L}}
\def\L{\mathbf{L}}

\def\Lbh{\widehat{\mathbf{L}}}
\def\Lbt{\widetilde{\mathbf{L}}}

\def\p{\mathbf p}

\def\R{\mathbf R}

\ifx\BlackBox\undefined
\newcommand{\BlackBox}{\rule{1.5ex}{1.5ex}}
\fi

\def\x{\mathbf x}

\def\y{\mathbf y}

\def\one{\mathbf 1}

\ifisneurips
   \def\st{\deff(\alpha\Lbh)}
   
\else
   \def\st{\tilde{s}}
   
\fi

\def\ee{\mathrm{e}}

\def\B{\mathbf B}

\def\A{\mathbf A}

\def\U{\mathbf U}

\def\St{\widetilde{S}}

\def\I{\mathbf I}

\def\A{\mathbf A}
\def\P{\mathbf P}

\def\E{\mathbb E}

\def\R{\mathbb R}

\def\ZZ{\mathbb Z}
\def\tr{\mathrm{tr}}

\newcommand{\defeq}{\stackrel{\textit{\tiny{def}}}{=}}

\let\origtop\top
\renewcommand\top{{\scriptscriptstyle{\origtop}}}

\definecolor{silver}{cmyk}{0,0,0,0.3}
\definecolor{yellow}{cmyk}{0,0,0.9,0.0}
\definecolor{reddishyellow}{cmyk}{0,0.22,1.0,0.0}
\definecolor{black}{cmyk}{0,0,0.0,1.0}
\definecolor{darkYellow}{cmyk}{0.2,0.4,1.0,0}
\definecolor{darkSilver}{cmyk}{0,0,0,0.1}
\definecolor{grey}{cmyk}{0,0,0,0.5}
\definecolor{darkgreen}{cmyk}{0.6,0,0.8,0}

\newcommand{\Green}[1]{{\color{darkgreen}  {#1}}}
\newcommand{\Blue}[1]{\color{blue}{#1}\color{black}}
\newcommand{\Brown}[1]{{\color{brown}{#1}\color{black}}}

\ifx\proof\undefined
\newenvironment{proof}{\par\noindent{\bf Proof\ }}{\hfill\BlackBox\\[2mm]}
\fi

\ifx\theorem\undefined
\newtheorem{theorem}{Theorem}
\fi

\ifx\example\undefined
\newtheorem{example}{Example}
\fi

\ifx\property\undefined
\newtheorem{property}{Property}
\fi

\ifx\lemma\undefined
\fi

\ifx\proposition\undefined
\fi

\ifx\remark\undefined
\newtheorem{remark}{Remark}
\fi

\ifx\corollary\undefined
\newtheorem{corollary}{Corollary}
\fi

\ifx\definition\undefined
\newtheorem{definition}{Definition}
\fi

\ifx\conjecture\undefined
\newtheorem{conjecture}[theorem]{Conjecture}
\fi

\ifx\axiom\undefined
\newtheorem{axiom}[theorem]{Axiom}
\fi

\ifx\claim\undefined
\newtheorem{claim}[theorem]{Claim}
\fi

\ifx\assumption\undefined
\newtheorem{assumption}{Assumption}
\fi

\def\ie{i.e.,\xspace}
\def\eg{e.g.,\xspace}

\def\nystrom{Nystr\"om\xspace}
\def\dppvfx{\textsc{DPP-VFX}\xspace}
\def\bless{\textsc{BLESS}\xspace}
\def\pbless{\textsc{BLESS-I}\xspace}
\def\alphadpp{\textsc{$\alpha$-DPP}\xspace}
\newcommand{\wh}[1]{\widehat{#1}}

\newcommand{\Ltrace}{\tr(\Lb)}
\newcommand{\tinit}{\text{init}}
\ifisneurips\fi
\newcommand{\kerfunc}{K}
\newcommand{\featmap}{\varphi}
\newcommand{\statespace}{\mathcal{X}}
\newcommand{\rkhs}{\mathcal{H}}

\newcommand{\wt}[1]{\widetilde{#1}}

\newcommand{\wb}[1]{\overline{#1}}
\newcommand{\deff}{d_{\textnormal{eff}}}
\newcommand{\edeff}{\wh{d}_{\textnormal{eff}}}
\newcommand{\adeff}{\wt{d}_{\textnormal{eff}}}

\newcommand{\coldict}{\mathcal{D}}
\newcommand{\indfunc}{\mathbb{I}}
\newcommand{\Poisson}{\text{Poisson}}
\newcommand{\Bernoulli}{\text{Bernoulli}}
\newcommand{\Multinomial}{\text{Multinomial}}
\newcommand{\Multinoulli}{\text{GenBinomial}}
\newcommand{\Binomial}{\text{Binomial}}
\newcommand{\Uniform}{\text{Uniform}}
\newcommand{\unifset}{\rho}
\newcommand\acO{\wt{\mathcal O}}
 \usepackage{thmtools,thm-restate}
\declaretheorem{lemma}
\declaretheorem{proposition}

\title{Sampling from a $k$-DPP without looking at all items}

\definecolor{redstar}{HTML}{B65353}

\author{
Daniele Calandriello\thanks{Equal contribution.}\\
DeepMind Paris\\
\texttt{dcalandriello@google.com}
\And
Micha{\l } Derezi\'{n}ski\textcolor{redstar}{\footnotemark[1]}\\
University of California, Berkeley\\
\texttt{mderezin@berkeley.edu}
\And
Michal Valko\\
DeepMind Paris\\
\texttt{valkom@deepmind.com}
}

\begin{document}

\maketitle

\begin{abstract}
Determinantal point processes (DPPs) are a useful probabilistic model
for selecting a small diverse subset out of a large collection of items,
with applications in summarization, stochastic
optimization, active learning and more. Given a kernel function and
a subset size $k$, our goal is to sample $k$ out of $n$ items with
probability proportional to the determinant of the kernel matrix
induced by the subset (a.k.a.~$k$-DPP). Existing $k$-DPP sampling
algorithms require an expensive preprocessing step which involves
multiple passes over all $n$ items, making it infeasible for large
datasets. A na\"ive heuristic addressing this problem is to uniformly
subsample a fraction of the data and perform $k$-DPP sampling only on
those items, however this method offers no guarantee that the
produced sample will even approximately resemble the target
distribution over the original dataset. In this paper, we develop an
algorithm which adaptively builds a sufficiently large uniform sample
of data that is then used to efficiently generate a smaller
set of $k$ items, while ensuring that this set is drawn exactly from the
target distribution defined on all $n$ items.
We show empirically that our
algorithm produces a $k$-DPP sample after observing only a small
fraction of all elements, leading to several orders of magnitude faster
performance compared to the state-of-the-art.

\end{abstract}

\section{Introduction}
Selecting $k$ diverse items out of a larger collection of $n$ items
is a classical problem in computer science which naturally emerges in many
tasks such as summarization (select $k$ phrases) and recommendation (select
$k$ articles/ads to show to the user). An increasingly popular approach
to model and quantify diversity in this
subset selection problem is that of determinantal point processes (DPPs).
Given a set $[n] \defeq \{1, \ldots, n\}$ of $n$ items and a target
size $k$, one can define a DPP of size $k$ (known as a $k$-DPP)
through an $n\times n$ posivite semi-definite (PSD) similarity matrix
$\L$ (also known as the kernel matrix).
The matrix $\L$ encodes the similarities between items, and the user must choose
it so that $[\L]_{ij}$ is larger the more the $i$-th and $j$-th items are similar.
Given $k$ and $\L$, we define $S \sim k$-$\DPP(\L)$
as a distribution over all ${n \choose k}$ index subsets
$S\subseteq [n]$ of size $k$, such that $\Pr(S) \propto \det(\L_S)$
is proportional to the determinant of the sub-matrix $\L_S$ induced by
the subset. DPPs have found numerous applications in machine
learning, not only for summarization \cite{dpp-summarization,dpp-video,dpp-salient-threads,pmlr-v80-celis18a}
and recommendation \cite{dpp-shopping,NIPS2018_7805}, but also in experimental
design \cite{bayesian-experimental-design,symmetric-polynomials},
stochastic optimization \cite{dpp-minibatch,randomized-newton},
Gaussian Process optimization \cite{NIPS2016_6452},
low-rank approximation
\cite{pca-volume-sampling,more-efficient-volume-sampling,nystrom-multiple-descent}, and
more (recent surveys include
\cite{dpp-ml,dpp-stats,dpp-randnla}). Note that early work on DPPs
focused on a \emph{random-size} variant, which we
denote $S\sim \DPP(\L)$, where the subset size is allowed to take any value
between $0$ and $n$, and the role of parameter $k$ is replaced by the expected size $\E[|S|]=
\deff(\L) \defeq \tr\left(\L(\L + \I)^{-1}\right)$.
The quantity $\deff(\L)$ is known in randomized linear algebra
\cite{ridge-leverage-scores,dpp-randnla} and learning theory
\cite{caponnetto2007optimal} as the effective dimension. While
random-size DPPs exhibit deep connections to many scientific domains
\cite{dpp-independence}, the \emph{fixed-size} $k$-DPPs are typically
more practical from a machine learning stand-point \cite{k-dpp}.

\textbf{Sampling from a $k$-DPP.} The first $k$-DPP samplers
scaled poorly, as they all relied on an eigendecomposition \cite{k-dpp}  of
$\L$ taking $\cO(n^3)$ time.
Replacing the eigendecomposition with a Cholesky factorization can increase numerical stability \cite{dpp-noeig},
and empirical performance \cite{poulson2019high-performance} thanks to
dynamically-scheduled, shared-memory parallelizations, but still
ultimately require $\cO(n^3)$ time. A number of methods have
been proposed which use approximate eigendecomposition
\cite{dpp-coreset,dpp-nystrom} to reduce the computational cost,
however these approaches provide limited guarantees on the accuracy of sampling.

To improve scalability, several approaches
based on Monte-Carlo sampling were introduced, using rejection or Gibbs sampling.
The fastest MCMC sampler for  $k$-DPPs, to the best of our knowledge, is
by \cite{rayleigh-mcmc} and has  $n\cdot\poly(k)$
complexity, i.e., asymptotically much faster than the cost of eigendecomposition.
However these MCMC methods do \emph{not} sample exactly from the $k$-DPP distribution,
and can only guarantee that the final sample will be close in distribution
to a $k$-DPP. Moreover these guarantees only hold \emph{after mixing}, which is difficult
to verify and requires at least $\cO(nk^2)$ time, making MCMC methods not applicable
when $n$ is large.

\begin{wrapfigure}{R}{.44\textwidth}
  \centering
\vspace{-2\baselineskip}
    \begin{tabular}{r|l}
      &Complexity
      \\
              \hline
      \cite{k-dpp,dpp-noeig,poulson2019high-performance,more-efficient-volume-sampling}
&$n^3$\\[1mm]
      \dppvfx \cite{derezinski2019exact}& $n\cdot k^{10} + k^{15}$\\[1mm]
$\alpha$-DPP (this paper)& $(\beta n\cdot k^6+k^9)\sqrt k$
    \end{tabular}
    \captionof{table}{\small Runtime comparison of exact $k$-DPP
      sampling algorithms. Here, $\beta\leq 1$ is the fraction of
      items observed by $\alpha$-DPP (see Theorem~\ref{thm:informal-main}).}
    \vspace{-3mm}
    \label{tab:comp}
\end{wrapfigure}

A recent line of works \cite{derezinski2019exact,dpp-intermediate},
using the ideas from \cite{leveraged-volume-sampling,correcting-bias},
developed sampling algorithms
specially designed for a \emph{random-size} DPP (as opposed to a $k$-DPP), which avoid
expensive decomposition of the kernel while sampling exactly from $S\sim\DPP(\L)$.
In particular, they showed that it is sufficient to first choose an intermediate subset
$\sigma \subseteq [n]$ sampled i.i.d.\,from the \emph{marginal}
distribution of the DPP,  \ie $\P(i \in \sigma) \approx \P(i \in S)$, and
then sample from a DPP restricted to the items indexed by
$\sigma$. Since the size of $\sigma$ is typically much less
than $n$, this leads to a more efficient algorithm.
Note that rescaling $\DPP(\L)$ into $\DPP(\alpha\L)$ using some constant~$\alpha$
only changes the expected size of $S$ from $\deff(\L)$ to $\deff(\alpha\L)$.
By accurately choosing an appropriate $\alpha_{\star}$, one can boost the probability that the random
size of $S$ is exactly $k$, and convert a DPP sampler into a $k$-DPP sampler
by repeatedly sampling $S \sim \DPP(\alpha_{\star}\L)$ until $S$ has
size~$k$. Based on this reduction, \citet{derezinski2019exact}
gave the first algorithm (\dppvfx) which is capable of exact sampling
from a $k$-DPP in time $n\cdot\poly(k)$. However,
when sampling from $k$-DPPs, the approach of
\cite{derezinski2019exact} has \emph{two major limitations}:
\begin{enumerate}[leftmargin=*]
  \item \dppvfx has an $\Omega(n)$ runtime bottleneck, since it requires
    computing all $n$ marginals, one for each item, in
    order to define the i.i.d.\/ distribution of $\sigma$, which may
    be infeasible for very large $n$.
\item
  The reduction used by \cite{derezinski2019exact} to convert a DPP
  sampler into a $k$-DPP sampler
   increases the time complexity by a factor
  of at least $k^4$, resulting in a $\acO(n\cdot k^{10}+k^{15})$ runtime.
\end{enumerate}

In this paper, we address both of these limitations by introducing a
new algorithm called $\alpha$-DPP, which 1) does not need to compute all
of the marginals, and 2) uses a new efficient reduction to convert
from a random-size DPP to a fixed-size $k$-DPP (see Table
\ref{tab:comp} for comparison).

\textbf{Main contribution: uniform intermediate sampling for $k$-DPPs.}
To resolve the $\Omega(n)$ runtime bottleneck, we use an additional
intermediate sample $\unifset$ based on \emph{uniform} sub-sampling.
Since uniform sampling can be implemented without looking at the actual items
in the collection, this means that we do not even have to look at any
item outside of $\unifset$.
The only necessary assumption required by our approach is that the maximum
entry (\ie similarity) of $\L$ is bounded by a constant $\kappa^2$. However,
to simplify exposition we also assume w.l.o.g.~that $\deff(\L) \geq k$ (see \Cref{s:binary-search}).

In particular, we 1) sample $\unifset$ uniformly out of $[n]$,
then 2) only approximate the marginal probabilities of items in $\unifset$
to compute $\sigma$, and finally 3) downsample $\sigma$ into
a DPP sample $S$. To guarantee that $S$ is distributed exactly according to the DPP
it is crucial that $\unifset$ is diverse enough. We show
that sampling a $k^2/\deff(\L)$ fraction of $[n]$ into $\unifset$ (\ie $|\unifset| \approx k^2/\deff(\L) \cdot n$)
is enough. Since all the expensive computation is performed only
on $\unifset$, this gives us a $\deff(\L)/k^2$ speedup
over existing methods.
\vspace{2mm}

\begin{theorem}\label{thm:informal-main}
Given any $\L \succeq \mathbf{0}$ with $\max_{ij} \L_{ij} \leq
\kappa^2$ and $1 \leq k \leq \deff(\L)$, there exists an algorithm that returns
$S \sim k$-$\DPP(\L)$, and with probability $1 - \delta$ runs in time
$$\acO\big((\beta n\cdot k^6+k^9)\sqrt k\log(1/\delta)\big),$$
where $\beta \leq\min\big\{k^2\kappa^2 /\deff(\L),1\big\}$ is the fraction
of items observed by the algorithm.
\end{theorem}
In the derivation of \Cref{thm:informal-main} we make several novel contributions.
First, we provide a DPP sampler that
given $\L$ and a rescaling $\alpha \leq 1$ leverages a mixture of uniform and rejection sampling
to sample from $\DPP(\alpha\L)$ observing only an $\alpha\kappa^2k$ fraction of
the items. We then show that the optimal rescaling $\alpha_{\star}$ required
by the reduction from $k$-DPP to DPP can be bounded with $\alpha_{\star} \leq \cO(k/\deff(\L))$,
and thus our rescaling-aware sampler can sample from $k$-DPPs looking
only at a $k^2/\deff(\L)$ fraction of the items. Finally, we provide
an efficient search algorithm to find a close approximation $\hat{\alpha}$
of $\alpha_{\star}$.

\textbf{Model misspecification and computational free lunch.}
Our result can be also interpreted from a perspective of model misspecification.
Note that every time the users define a $k$-DPP they
also implicitly define a random size $\DPP(\L)$.
Moreover, the natural expected sample size (\ie implicit number of unique items in $[n]$)
of $\DPP(\L)$ is $\deff(\L)$,
which does not depend on the desired size~$k$.
Therefore, if $\L$ is not chosen appropriately
$\deff(\L)$ might be much larger than $k$, and the $k$-DPP is selecting
$k$ unique items out of a much larger implicit pool of $\deff(\L) \gg k$ unique items.
In this case, it is possible to consider only a small $k^2/\deff(\L)$ fraction of the items
selected uniformly at random and still have enough unique items to sample a diverse $k$-subset.
Our result shows for the first time that it is possible to take advantage of this modeling disagreement between $k$ and $\deff(\L)$ to gain computational
savings while still sampling \emph{exactly} from the DPP, \ie a computational free lunch.

\textbf{Binary search reduction from k-DPP to DPP.} Both our approach and the one of \citet{derezinski2019exact} rely on first implementing
an efficient random-size DPP sampler, followed by the usage of a black-box construction
based on rejection sampling to transform the DPP sampler into a $k$-DPP sampler.
However the reduction of \citet{derezinski2019exact} requires access to a
high-precision estimate of $\deff(\alpha\L)$ in order to appropriately
tune $\alpha$. This makes optimizing $\alpha$ the bottleneck in the reduction from $k$-DPP to DPP,
and therefore there is a large computational gap between the two problems.
We close this gap thanks to a novel approach to find a suitable rescaling $\alpha$
based not on optimization but rather on binary search. Crucially, to find a suitable $\alpha$
this approach does not require an estimate of $\deff(\alpha\L)$, but only $\cO(\sqrt{k}\log(n))$ black-box calls
to a DPP sampler. Therefore, it can transform any random size DPP sampler
into a $k$-DPP sampler with only a $\sqrt{k}$ overhead,
and could be applied to any future improved sampler beyond this paper.

\section{Sampling from a rescaled DPP with intermediate uniform subsampling}
\label{s:alpha-sampler}
In this section we focus on a specific class of DPPs, $S \sim \DPP(\alpha\L)$,
specified using a rescaling $\alpha \leq 1$
and a similarity matrix $\L$, which we refer to as rescaled DPPs. The main result
of the section is showing that a sufficiently large subset selected uniformly
at random can be used as an intermediate sample to sample from a
rescaled DPP without looking at all of the items.
The main reason to focus on rescaled DPPs is because they naturally appear when
reducing $k$-DPP sampling to DPP sampling, where rescaling is used to align
the random size of the DPP and $k$. This is going to be the focus of the next section.
However the approach proposed in this section
is not limited to rescaled DPPs, but under the right assumptions can be extended
to accelerate sampling from generic DPPs. We will discuss these extensions
at the end of the section.\vspace{-1mm}
\paragraph{Notation}
We use $[n]$ to denote the set $\{1,\dots,n\}$. For a matrix
$\B\in\R^{n\times m}$ and index sets $C$, $D$, we use $\B_{C,D}$
to denote the submatrix of $\B$ consisting of the intersection of rows
indexed by $C$ with columns indexed by $D$.  If $C=D$, we use a
shorthand $\B_{C}$ and if $D=[m]$, we may write $\B_{C,[m]}$.
Finally, we also allow $C,D$ to be multisets or
sequences, in which case each row/column is duplicated in the matrix
according to its multiplicity (and in the case of sequences, we
order the rows/columns as they appear in the sequence).
Note that with this notation if $\L=\B\B^\top$ then
$\L_{C,D}=\B_{C,[n]}\B_{D,[n]}^\top$.

\vspace{-2mm}
\subsection{Background: distortion-free intermediate sampling.}\label{subs:background}
Rather than sampling directly from the target DPP, intermediate
sampling \cite{dpp-intermediate,leveraged-volume-sampling}
first selects an intermediate subset $\sigma$ from $[n]$, and then refines it
by extracting $S$ from $\sigma$. Crucially, if $\sigma$ is selected according
to a so-called Regularized DPP ($\Vol$), this is equivalent to sampling $S$ from a DPP.
\begin{definition}\label{d:r-dpp}
  For any psd matrix $\L \in \R^{n\times n}$, distribution
  $p\defeq \{p_i\}_{i=1}^n$ and $r>0$, define $\Lbt \in \R^{n\times n}$ with~$\Lbt_{i,j} \defeq \frac{\L_{i,j}}{r\sqrt{p_ip_j}}$. We define an $\Vol_p^r(\L)$
  as distribution over events $A\subseteq\bigcup_{k=0}^\infty[n]^k$ such that\vspace{-2mm}
  \begin{align*}
    \Pr(A) \defeq\E_{\sigma}\big[\one_{[\sigma\in A]}\det(\I +
    \Lbt_\sigma)\big]/{\det(\I + \L)}\CommaBin\quad\text{for}
    \quad\sigma=(\sigma_1,\dots,\sigma_t)\simiid p,\quad
    t\sim\mathrm{Poisson}(r).
  \end{align*}
\end{definition}
\begin{proposition}[{\citealp[Theorem 8]{dpp-intermediate}}]\label{t:composition}
  For any $\L$, $p$, $r,$ and $\Lbt$ defined as in Definition
  \ref{d:r-dpp},\vspace{-1mm}
  \begin{align*}
    \text{if}\quad \sigma\sim\Vol_p^r(\L)\quad\text{and}\quad
    S\sim\DPP(\Lbt_\sigma)\quad\text{then}\quad
    \{\sigma_i:i\!\in\! S\}\sim\DPP(\L).
  \end{align*}
\end{proposition}
A computationally inefficient but conceptually simple approach to rejection sample $\sigma$ is the following:\\*
1) compute all marginals $\P(i \in S) = \ell_i(\L) \defeq [\L(\I+\L)^{-1}]_{i}$ and sum to $\sum_{i=1}^n\ell_i(\L) = \deff(\L)$ \cite{ridge-leverage-scores};\\*
2) sample $t \sim \Poisson(c)$ and $\sigma \sim \Multinomial\left(t, \frac{\ell_1(\L)}{\deff(\L)}, \ldots, \frac{\ell_n(\L)}{\deff(\L)}\right)$ for an appropriate constant $c$;\\*
3) accept $\sigma$ w.p.~$\tfrac{\det(\I + \wt{\L}_{\sigma})}{C\det(\I + \L)}$, where $C$ is an appropriate constant used to make the rejection step valid.

All existing intermediate sampling algorithms \cite{dpp-intermediate,derezinski2019exact,leveraged-volume-sampling,correcting-bias} rely on this approach,
refining it to make use of efficient approximations of the marginals
$\ell_i(\L)$ and adapting the constants $c$ and $C$ to the data. However they all share a common bottleneck:
to sample $\sigma$ i.i.d.~they need to approximate all marginals $\ell_i(\L)$
and the normalization constant $\deff(\L)$, and therefore the final
runtime scales as $n\cdot\poly(k)$.
While this is much smaller than the $\cO(n^3)$ required by an exact sampler,
it still becomes quickly unfeasible when $n$ is very large. In what follows
we will introduce another approach to sample from an $\Vol$ that does not
require to approximate the marginals of all items, but only the items selected
in a preliminary \emph{uniform} intermediate sample.

\subsection{Faster DPP sampling with uniform intermediate sampling}

\begin{wrapfigure}[21]{R}{.45\textwidth}
   \vspace{-2\baselineskip}
 \begin{minipage}{1.02\linewidth}
 \begin{algorithm}[H]
   \caption{\alphadpp sampler}
 \label{alg:alpha-dpp-sampler}
 \begin{algorithmic}[1]
 \renewcommand{\algorithmicrequire}{\textbf{Input:}}
 \REQUIRE $\alpha$, $\L$, $\coldict$, $\W$, $r \geq 1$
\STATE Set $\Lbh = \W^{1/2}\L_{\coldict,\coldict}\W^{1/2} \in \R^{m \times m}$
\REPEAT
\STATE Sample $u \sim \Poisson(re^{1/r}\alpha n\kappa^2)$
\STATE Sample $\unifset = \Uniform(u, [n])$
\FOR{$j = \{1, \dots, u\}$}
\STATE Compute $l_{\unifset_j}$ using Eq.~(\ref{eq:rls-approx})
\STATE
Sample $z_j \sim \Bernoulli(l_{\unifset_j}/(\alpha\kappa^2))$\label{line:reject-binomial}\hspace*{-1cm}
\ENDFOR
\STATE Set $\sigma = \{\unifset_j: z_j = 1\}$, $t = |\sigma|$
\STATE Set $[\Lbt_{\sigma}]_{ij} = \tfrac{1}{r\sqrt{l_{\sigma_i} l_{\sigma_j}}}[\L]_{\sigma_i\sigma_j}$
\STATE $\textit{Acc}\sim\!
   \Bernoulli\Big(\tfrac{\ee^{\st}\det(\I+ \alpha\Lbt_\sigma)}{\ee^{t/r}\det(\I+\alpha\Lbh)}
 \Big)$ \label{line:acc}
 \UNTIL{$\textit{Acc}=\text{true}$\label{line:rep2}}
 \STATE Sample $\St\sim \DPP\big(\alpha\Lbt_\sigma\big)$\label{line:sub}
   \RETURN $S = \{\sigma_i: i\!\in\! \St\}$
 \end{algorithmic}
 \end{algorithm}
\end{minipage}
\end{wrapfigure}
We now introduce our novel $\alpha$-rescaled DPP sampler, called
\alphadpp (see \Cref{alg:alpha-dpp-sampler}).
It requires as input a rescaling $\alpha$, a similarity matrix $\L$ and a parameter
$r$ that will be used to tune the Poisson sampling step of \Cref{t:composition}
approach.
It also requires as input a \emph{dictionary} $\coldict$ containing $m$ elements,
and set of weights stored in a diagonal matrix $\W \in \R^{m \times m}$.
A dictionary is a subset of items $\coldict \subseteq [n]$
such that reweighting the items in $\coldict$ by $\W$ provides a good approximation
of $\L$, so that the approximate marginals
\begin{align}\label{eq:rls-approx}
l_i \defeq \alpha[\L - \alpha\L_{[n],\coldict}^\transp(\alpha\L_{\coldict} + \W^{-1})^{-1}\L_{[n],\coldict}]_{i}
\vspace{-.25\baselineskip}
\end{align}
computed using $\coldict$ and $\W$ are close to the true marginals $\ell_i$ (see \Cref{a:misc-proofs}).
Compared to the meta-approach of \Cref{t:composition}, the main technical difference
is that rather than sampling directly $t$ from an appropriate Poisson,
and then $\sigma$ from a Multinomial, we introduce an intermediate uniform sampling
step. In particular, we first sample a Poisson $u$, and then uniformly sample
a subset $\unifset$ containing $u$ items. We then compute an approximation
$l_i$ of the marginal $\ell_i$ only for the items in $\unifset$,
and downsample $\unifset$ into $\sigma$ using rejection sampling (\Cref{line:reject-binomial}).
Finally, we accept or reject $\sigma$ (\Cref{line:acc})
and then downsample $\sigma$ into $S$ using a standard DPP sampler on
the smaller $\wt{\L}_{\sigma}$.

\Cref{alg:alpha-dpp-sampler} is not simply a different
implementation of the approach of \Cref{t:composition}, since even if Multinomial sampling is implemented with lazy evaluations of $l_i$,
we would still need to compute the \emph{normalization} constant of the Multinomial,
which strictly requires computing all $l_i$. Similarly, the rejection test of \Cref{line:acc} is also designed to accept as many candidates as possible without requiring the computation of the normalization constant as in \cite{derezinski2019exact}. Rather
our approach is a novel method to sample from an $\Vol$
using Poisson rejection sampling. In particular, we prove not only that
it always returns an $S$ sampled according to the exact DPP distribution,
but also that if the dictionary satisfies certain conditions,
the main of which is $(\varepsilon,\alpha)$-accuracy (see \Cref{a:misc-proofs} and \cite{calandriello_disqueak_2017}),
then the algorithm will generate $S$ quickly.
\begin{theorem}\label{thm:alpha-sampling}
Given any $\L \succeq \mathbf{0}$, dictionary $\coldict$, $\W\succ \mathbf{0}$, $r \geq 1$ and $\alpha > 0$, \alphadpp returns
$S \sim \DPP(\alpha\L)$. Moreover, if $r \geq \deff(\alpha\L) \geq 1/2$, $\coldict$ and $\W$ are $(1/\deff(\alpha\L), \alpha)$-accurate,
$\coldict$ satisfies $|\coldict| \leq 10\deff(\alpha\L)$, and $\deff(\alpha\Lbh) \leq 10\deff(\alpha\L)$, w.p.~$1 - \delta$ \alphadpp runs in time\vspace{-1mm}
\begin{align*}
  \cO\Big(\left[ \min\{\alpha\kappa^2\deff(\alpha\L), 1\}\cdot n \cdot
  \deff(\alpha\L)^6\log^2(n/\delta) +
  \deff(\alpha\L)^9\log^3(n/\delta)\right ] \cdot \log(1/\delta)\Big).
  \end{align*}
\end{theorem}
\vspace{-1mm}

The main implication of our result is that the intermediate distribution based
on marginals can be replaced more and more accurately with a uniform distribution
as $\alpha$ becomes smaller. This results in having to compute marginals
only for a $\min\{\alpha\kappa^2\deff(\alpha\L), 1\}$ fraction of the $n$ items.
This speedup can be significant when the rescaling $\alpha$ is very small,
as is the case when we want to sample a small number of items out of a large
collection.
Compared to other exact DPP samplers, such as \dppvfx,
our \alphadpp is strictly faster by roughly a $1/(\alpha\kappa^2\deff(\alpha\L))$ factor
when implemented with an appropriate caching strategy for the
estimates $l_i$ (see \Cref{a:misc-proofs}).
Further, unlike MCMC samplers, \alphadpp is an \emph{exact} sampler. Moreover, there is no
known MCMC approach that can achieve a runtime sub-linear in $n$ when $\alpha$
is small as \alphadpp.

An $(\varepsilon, \alpha)$-accurate dictionary
that also satisfies the other conditions can be generated using
a slight modification of the \bless algorithm \cite{NIPS2018_7810}, that we call \pbless algorithm, presented
in \Cref{a:poisson-bless}.
However, note that since the marginals $\ell_i$ are equivalent to
the ridge leverage scores \cite{ridge-leverage-scores} of item $i$,
we can replace \pbless with any present or future algorithm for leverage score sampling that can
be modified to be rescaling-aware \cite{calandriello_disqueak_2017,NIPS2018_7810}.
Moreover, note that \pbless also returns an estimate
of $\deff(\alpha\L)$ that is sufficiently accurate to tune $r$ and $\varepsilon$.
At the same time, our analysis could be excessively conservative, and instead of trying
to set $r$ and $\varepsilon$ using $\deff(\alpha\L)$ as suggested by \Cref{thm:alpha-sampling},
a more practical strategy is to start with a constant
$r$ and increase it slowly if the sampler is rejecting with a too low probability,
using a doubling schedule to preserve overall time complexity.

\textbf{Proof sketch.} The proof is divided in two parts, proving that \alphadpp is an exact sampler
(\Cref{lem:a-dpp-exact}) and that under the right conditions it is efficient (\Cref{lem:a-dpp-efficient}).

For the first part we once again rely on the approach of \Cref{t:composition},
but with the added difficulty of not being allowed to compute all the marginals.
To avoid this bottleneck, we show that:\vspace{-1mm}
\begin{enumerate}
\item[A)] sampling $t \sim \Poisson(r)$ and $\sigma \sim
  \Multinomial(t, \{\ell_i/\deff(\alpha\L)\}_{i=1}^n)$;\quad and
\item[B)] sampling $n$ independent $s_i \sim \Poisson(r'\ell_i(\alpha\L))$, and adding $s_i$ copies of item $i$ to $\sigma$,
\end{enumerate}\vspace{-1mm}
are equivalent for an appropriate choice of $r$ and $r'$, \ie
we prove that the $\sigma$ generated by both approach A and B follow the same distribution.
However, unlike approach A, approach B does not require computing
a normalization constant, \ie it samples from \emph{unnormalized} probabilities.
Moreover, if we know an upper bound on the marginals we can further reduce the number of marginals that need
to be computed. In our case we use
the bound $\ell_i \leq \alpha\kappa^2$, and show that
\vspace{-1mm}
\begin{enumerate}
  \item[C)] sampling $n$ Poisson independently $u_i \sim
\Poisson(r'\alpha\kappa^2)$, only if $u_i > 0$ computing
$\ell_i$ and sampling $s_i \sim \Binomial(u_i,
\ell_i/(\alpha\kappa^2))$, and adding $s_i$ copies of item $i$ to
$\sigma$
\end{enumerate}\vspace{-1mm}
once again generates $\sigma$ strictly equivalent to the ones of approach B and A.
The added advantage of approach C over the others is that only the marginals of items with $u_i > 0$
are actually computed, and there is no need to compute a normalization constant.
Starting from this new approach, to obtain our \alphadpp sampler (\Cref{alg:alpha-dpp-sampler})
we simply replace the $n$ Poisson $u_i \sim \Poisson(r'\alpha\kappa^2)$ with a single
$u \sim \Poisson(r'\alpha\kappa^2 n)$ followed by uniform sampling,
and replace the exact $\ell_i$ with approximate $l_i$.

For the second part we derive a lower bound on the acceptance probability
similar to the one from \citet{derezinski2019exact}. However, while they
use an $n \times \deff(\alpha\L)$ \nystrom approximation of the matrix $\L$,
to avoid direct dependencies on $n$ we are forced to use a less stable
approximation $\Lbh$. As a result, controlling $\deff(\alpha\Lbh)$
requires a more careful analysis.

\textbf{Beyond uniform subsampling.} One of the implications of our analysis
is that more adaptive upper bounds on the marginals $\ell_i$ could further
speedup our \alphadpp sampling approach.
In particular, we chose uniform sampling, \ie a uniform upper bound,
for its conceptual simplicity and because knowing an upper bound $\kappa^2$
on the entries of the similarity matrix usually does not require looking at
the items, \eg $\kappa^2$ is always equal to 1 for Gaussian similarity, Cosine similarity or other
self normalized similarities. However for other similarities, such as linear
similarity, this bound could be very loose. A simple replacement is using
the actual diagonal of $\L$, which requires to look at all items and $\cO(n)$
time to compute but is usually very scalable. Ideally, one could imagine
designing a sequence of upper bounds starting from cheaper to more computationally
expensive, where more advanced techniques such as random projection are used
near the end to further filter candidate items.
 \section{Efficient reduction from k-DPP to rescaled DPP via binary search}
\label{s:binary-search}

Given our fast DPP sampler, we can see a $k$-DPP as a sampling process where we first sample
$S \sim \DPP(\alpha\L)$, check if the sample size $|S|$ is equal to $k$, and then
accept or reject the sample accordingly. Rescaling
$\L$ by a constant factor $\alpha$ only changes the expected size
$\deff(\alpha\L)$ (and not the $k$-DPP), with
$\alpha > 1$ increasing the expected size and $\alpha < 1$ decreasing it.
Thus, it is natural to imagine that there exists some $\alpha_\star$ for which
the acceptance probability is high. Indeed this was recently proven to be possible.
\citet{derezinski2019exact} show that if the \emph{mode}
$m_{\alpha_{\star}}$ of $S_{\alpha_\star}\sim\DPP(\alpha_\star \L)$ is equal to
$k$, then we will accept with probability at least $\Omega(1/\sqrt{k})$.
They also provide an
algorithm to find such an $\alpha_{\star}$. However,
this algorithm has a prohibitively high computational cost,
$\acO(nk^{10} + k^{15})$, because ensuring that the mode of
$\DPP(\alpha_\star\L)$ is \emph{exactly} $k$ requires an extremely
accurate approximation of $\L$. Instead, our approach is to run a binary search
to find a good rescaling $\alpha$, which will terminate once the
acceptance probability is high enough, regardless of whether $k$ is
exactly the mode. Crucially, this binary search only
requires a black box $\DPP(\alpha\L)$ sampler (such as
our \alphadpp), and it only queries the sampler $\acO(\sqrt k)$ many
times. To prove that the binary search finds a good $\alpha$ in a
small number of steps, we establish a new property (Lemma~\ref{l:pb-new}) of the Poisson
Binomial distribution (the distribution of the subset sizes of
$\DPP(\alpha\L)$), which should be of independent interest.

\subsection{Binary search}
Our main result in this subsection is \Cref{alg:binary-search}, which requires only oracle
access to the samples from a random-size DPP, and finds a rescaling
$\hat\alpha$ which enables efficient rejection sampling from a
$k$-DPP. Note that the provided oracle sampler does not have to be
our \alphadpp sampler, so the algorithm could be paired with other samplers.

\begin{lemma}[{restate=[name=restated]binarysearch}]\label{l:binary-search}
  Suppose that we are given an integer $k$, a range
  $I=[\alpha_{\min},\alpha_{\max}]$ where
  $\alpha_{\max}=\gamma\alpha_{\min}$, and access
  to an oracle which, for any
  $\alpha\in I$, returns $S\sim
  \DPP(\alpha \L)$. If there exists $\alpha_\star\in I$ such that $k$
  is the mode of $|S|$ for $S\sim\DPP(\alpha_\star \L)$, then using
  $O\big(\sqrt k\log^2(k\log(\gamma)/\delta)\big)$
  calls to the oracle we can find $\hat\alpha\in I$ such that with
  probability $1-\delta$ we have
  \begin{align*}
    \Pr(|S| = k) = \Omega\big(\tfrac1{\sqrt k}\big),\qquad\text{for}\quad
    S\sim\DPP(\hat\alpha \L).
  \end{align*}
\end{lemma}\vspace{-2mm}
The distribution of subset size $|S|$ for $S\sim\DPP(\L)$
can be defined via the eigenvalues $\lambda_1\geq\lambda_2\geq ...$ of
$\L$ (see \cite{dpp-independence}): if we let $b_i\sim
\mathrm{Bernoulli}(\frac{\lambda_i}{\lambda_i+1})$ for $i\geq 1$, then
$\sum_i b_i$ is distributed identically to $|S|$. This distribution is known as the \emph{Poisson
  Binomial}, and it has been extensively studied in the probability
literature \cite{pb-review}. The recent result of
\cite{derezinski2019exact} on the probability of the mode of a Poisson
Binomial shows that it is possible to find $\hat\alpha$ satisfying the
condition of Lemma \ref{l:binary-search}.
\begin{lemma}[{restate=[name=restated]pbmode}]\label{l:pb-mode}
   There is an absolute constant $0<c<1$ such that for any Poisson
   Binomial distribution $p:\ZZ_{\geq 0}\rightarrow\R_{\geq 0}$,
with mode $k^*$ we have $p(k^*)\geq \frac{c}{\sqrt{k^*\!+1}}\cdot$
\end{lemma}\vspace{-2mm}
This result, however, does not provide an efficient way of
finding an $\hat\alpha$ such that the mode of the subset size distribution of
$\DPP(\hat\alpha\L)$ is $k$. We circumvent this problem by performing
a binary search (\Cref{alg:binary-search}) that looks for such an
$\hat\alpha$, but stops early when it reaches a
sufficiently good candidate, avoiding excess computations. To make this rigorous, we establish the
following new property of the Poisson Binomial distribution, which
should be of independent interest.
\begin{lemma}[{restate=[name=restated]pbnew}]\label{l:pb-new}
  Let $p:\ZZ_{\geq 0}\rightarrow\R_{\geq 0}$ be a Poisson Binomial distribution,
  and let $k\geq 1$ satisfy
$p(k)<\frac{c}{12\sqrt{3(k+1)}}\CommaBin$ where $c$ comes from Lemma
\ref{l:pb-mode}. Then, $P_{<k}=\sum_{i<k}p(k)$ and
$P_{>k}=\sum_{i>k}p(k)$ satisfy:
\begin{enumerate}[leftmargin=1cm]
  \vspace{-3mm}
\item if the mode of $p$ is less than $k$, then
$P_{>k}\leq \frac12 - \frac c{12}$; \vspace{-2mm}
\item if the mode of $p$ is greater than $k$, then $P_{<k}\leq \frac12
  - \frac{c}{12}\cdot$
\end{enumerate}
\end{lemma}\vspace{-1mm}
Informally, the above result states the following:
For any $k$, either its probability under the given Poisson Binomial
is at least $\Omega(\frac1{\sqrt k})$, or this $k$ splits the probability mass into two
uneven parts, with the larger one containing the mode. Thus, as long
as our candidate $\alpha$ does not yield high acceptance probability
for $k$, it is easy to make the branching decision in the binary search by estimating the
quantities $P_{>k}$ and $P_{<k}$ simply by repeated sampling from
$\DPP(\alpha\L)$. Note that if the condition on $p(k)$ is not
satisfied, then performing the branching decision could be very
expensive, but our algorithm avoids this possibility. The proof of
\Cref{l:binary-search} (\Cref{a:bin-search}) follows from Lemmas
\ref{l:pb-mode} and \ref{l:pb-new}.

\setlength{\textfloatsep}{5pt}
\begin{algorithm}[t]
  \caption{Binary search for initializing the $k$-$\DPP(\L)$ sampler}
\label{alg:binary-search}
{
\setstretch{1.4}
\begin{algorithmic}[1]
  \renewcommand{\algorithmicrequire}{\textbf{Input:}}
  \renewcommand{\algorithmicensure}{\textbf{Output:}}
\REQUIRE $0<\alpha_{\min}<\alpha_{\max}$, sampling oracle for
$\DPP(\alpha \L)$, integer $k$ and constants $C>0, \delta\in(0,1)$
\ENSURE $\hat\alpha$ such that $\DPP(\hat\alpha \L)$ can be used
to efficiently sample $k$-$\DPP(\L)$
\FOR{$s = \{1, \ldots, \lceil\log(\gamma)\rceil\}$}
\STATE \textbf{if} {$\alpha_{\max}/\alpha_{\min} < (1+\frac1{(k+3)^2})$\label{line:if1}} \textbf{then return} $\hat\alpha = \alpha_{\min}$
\STATE $\bar\alpha \leftarrow\sqrt{\alpha_{\min}\alpha_{\max}}$
\STATE Sample $S_1,...,S_t\simiid \DPP(\bar\alpha \L)$\quad
where\quad $t = C\sqrt k\log(s/\delta)$
\STATE $\hat P_k \leftarrow \frac1t \sum_{i=1}^t\one_{[|S_i|=k]}$
\STATE \textbf{if} {$\hat P_k\geq \frac12\cdot \frac{c}{12\sqrt{3(k+1)}}$\label{line:if2}} \textbf{then return}
$\hat\alpha = \bar\alpha$
\STATE  $(\hat P_{<k}, \hat P_{>k}) \leftarrow \big(\frac1t
\sum_{i=1}^t\one_{[|S_i|<k]},\ \frac1t
\sum_{i=1}^t\one_{[|S_i|>k]}\big)$

\vspace{.15cm}
\STATE \textbf{if} $\hat P_{<k}> \hat P_{>k}$ \textbf{then} $(\alpha_{\min},\alpha_{\max}) = (\bar\alpha,\alpha_{\max})$ \textbf{else} $(\alpha_{\min},\alpha_{\max}) = (\alpha_{\min},\bar\alpha)$
\label{line:branch}
\ENDFOR
\end{algorithmic}
}
\end{algorithm}

\subsection{Constructing the initial interval}

To initiate our binary search, we must first find a range of values
$[\alpha_{\min},\alpha_{\max}]$, which contains the desired $\alpha_\star$, and also
construct a sampling oracle for $\DPP(\alpha\L)$.
The binary search procedure is deliberately presented in a way that is agnostic to
how these two steps are accomplished, because a number of existing DPP
samplers could be adapted to take advantage of
\Cref{alg:binary-search}, including
\cite{dpp-noeig,poulson2019high-performance,dpp-intermediate,derezinski2019exact}.
Our implementation of these two steps is different than these previous
approaches in that it takes advantage of the structure of the kernel
so that it only has to look at a potentially small fraction of the
data points. We achieve this with a modified version
of the BLESS algorithm \cite{NIPS2018_7810}.

\begin{lemma}[{restate=[name=restated]algovalidinterval}]\label{lem:algo-valid-interval}
  W.p.~$1-\delta$ \pbless runs in time
 $\acO\left(\min\{\alpha_{\max}\kappa^2, 1\}nk^6 + k^9\right)$
  and satisfies:
  \vspace{-.5\baselineskip}
  {
  \setlength{\itemsep}{-2pt}
  \begin{enumerate}[leftmargin=1cm]
    \item The interval $[\alpha_{\min},\alpha_{\max}]$ is bounded by $\frac14(k-1)/\Ltrace \leq \alpha_{\min} \leq \alpha_{\max} \leq 8(k + 2)/\deff(\L)$
  \item    There is $\alpha_\star\in[\alpha_{\min},\alpha_{\max}]$
    for which $k$ is the mode of $|S|$ where
    $S\sim\DPP(\alpha_\star\L)$;
    \item The dictionary $\coldict^{\alpha_{\max}}$ satisfies the
      conditions from \Cref{thm:alpha-sampling} for any $\alpha\in[\alpha_{\min},\alpha_{\max}]$.
  \end{enumerate}
  }
\end{lemma}

The first two parts of the lemma ensure that the interval
$I=[\alpha_{\min},\alpha_{\max}]$ is a valid input for
the binary search in \Cref{alg:binary-search} and that its size
$\gamma = \alpha_{\max}/\alpha_{\min} \leq 4\Ltrace/\deff(\L)$
is bounded in the log-scale. The last part implies that
$\alpha$-DPP can be used by that algorithm as the oracle sampler.

At a high level, \Cref{alg:adjusted-bless} proceeds by starting with a
small $\alpha^0$ that is guaranteed to be a valid lower bound for the interval,
and for which a dictionary $\coldict^0$ can be constructed simply via
uniform sampling. Then we repeatedly double the $\alpha$ and refine
the dictionary, until we reach $\alpha^i$ such that we can ensure that
with high probability $\deff(\alpha^i\L) \geq k + 1$ which makes it
a valid upper bound for the interval (then, this $\alpha^i$
becomes $\alpha_{\max}$).

\subsection{Overall time complexity of $k$-DPP sampling}
Putting together all the results from the previous sections, we can
finally bound the computational complexity of our $k$-DPP sampler,
which first uses \pbless (\Cref{alg:adjusted-bless}) to construct a dictionary and search interval,
and then applies the binary search of (\Cref{alg:binary-search}) using
our $\alpha$-DPP sampler (\Cref{alg:alpha-dpp-sampler}) as the sampling oracle.
Once again note that in the following computational analysis we will use conservative
values for many parameters, notably $r$ from \alphadpp and $q$ from \pbless,
as they are suggested from the theory. However in practice it is always better to
start from a more optimistic value, and keep doubling them only if the sampler
repeatedly fails to accept. Importantly, samples generated this way will still
be exactly distributed according to the DPP, as all the approximations used
in our approach only influence the runtime of our algorithm, and not the correctness of
its acceptance, which always holds.

By \Cref{lem:algo-valid-interval}, the preprocessing step of running
\pbless takes $\acO\left(\min\{\alpha_{\max}\kappa^2,1\}nk^6 + k^9\right)$
and generates a dictionary $\coldict$ with size $\acO(k^3)$.
Since $\deff(\alpha\L) \leq \deff(\alpha_{\max}\L) \leq \cO(k)$ for all
$\alpha$ in the search interval, each call to the \alphadpp sampler also requires
at most $\acO\left(\min\{\alpha_{\max}\kappa^2k, 1\}nk^6 + k^9\right)$.
Finally, the binary search invokes $\alpha$-DPP at most $\acO(\sqrt{k})$ times
so the overall runtime is
$\acO\big((\min\{\alpha_{\max}\kappa^2k, 1\}nk^{6} +
    k^{9})\cdot\sqrt{k}\big)$.
We now provide a bound on $\alpha_{\max}$.
\begin{lemma}[{restate=[name=restated]lambdalowerbound}]\label{lem:lambda-lower-bound}
For any matrix $\Lb$ and $0 < \alpha \leq 1$, we have
$\deff(\alpha\Lb)/\deff(\Lb) \geq \alpha \geq \deff(\alpha\Lb)/\Ltrace$.
\end{lemma}
Applied to $\alpha_{\max}$, we obtain $\alpha_{\max} \leq \deff(\alpha_{\max}\L)/\deff(\L) \leq
\cO(k/\deff(\L))$, giving us the final runtime of
$\acO\big((\min\{k^2\kappa^2/\deff(\L), 1\}nk^{6} + k^{9})\cdot\sqrt{k}\big)$
reported in \Cref{thm:informal-main}.

\noindent\begin{minipage}{0.49\textwidth}
\centering
\includegraphics[width=.95\textwidth]{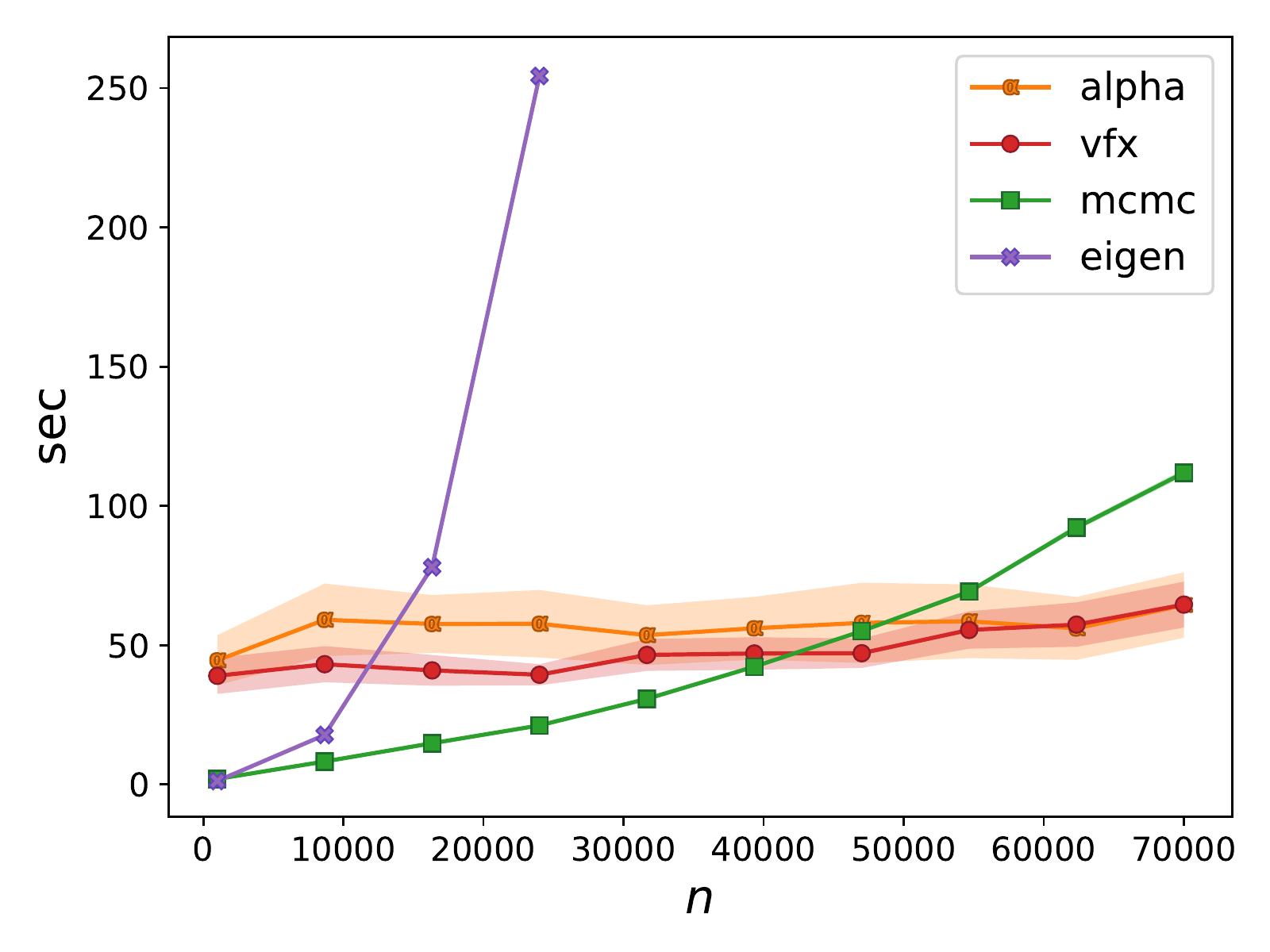}
\vspace{-3mm}
\captionof{figure}{Small scale experiment}\label{fig:small-scale}
\vspace{-2mm}
\end{minipage}\hfill
\begin{minipage}{0.49\textwidth}
\centering
\includegraphics[width=.95\textwidth]{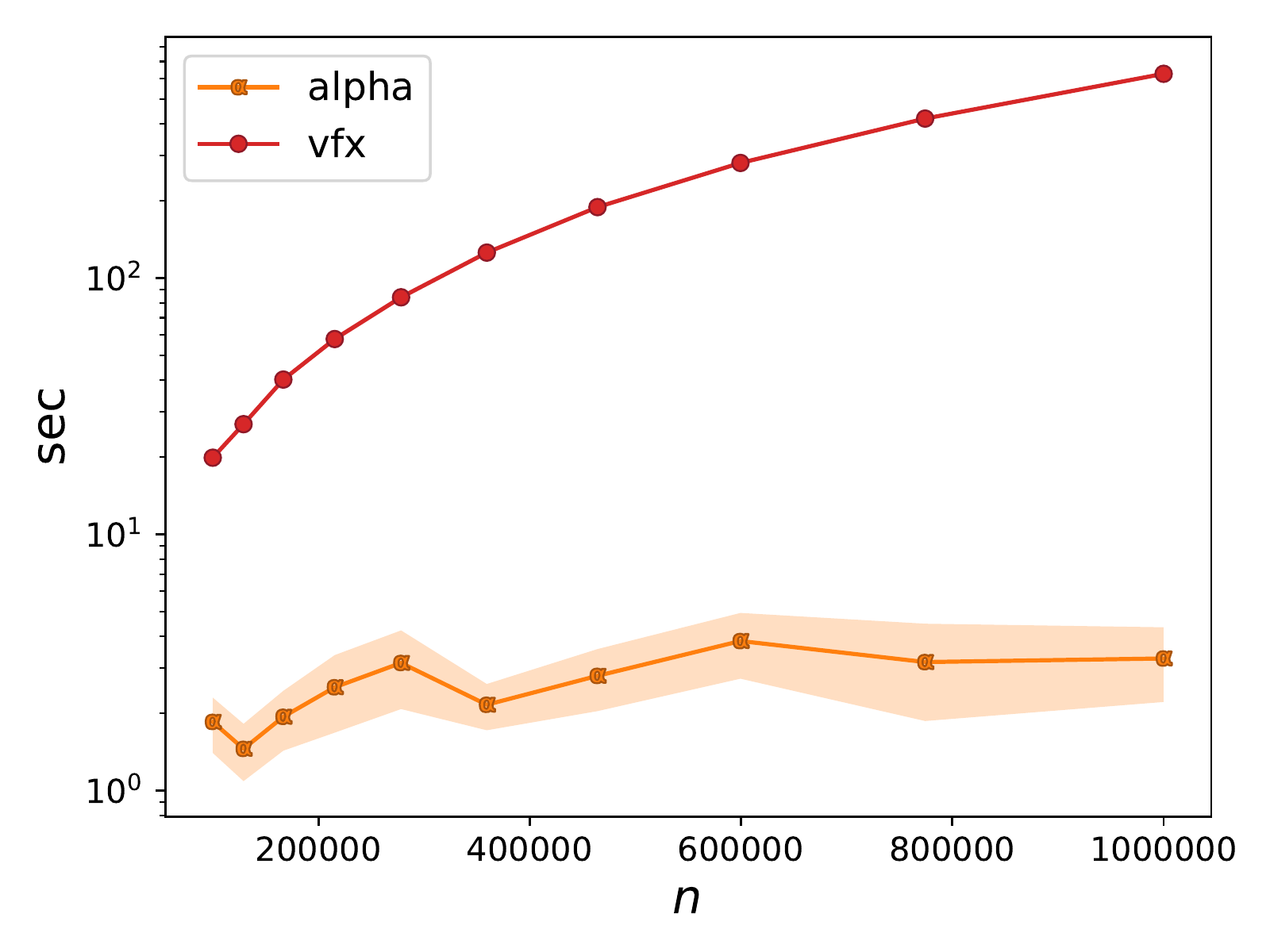}
\vspace{-3mm}
\captionof{figure}{Large scale experiment.}\label{fig:large-scale}
\vspace{-2mm}
\end{minipage}
\section{Experiments}
\vspace{-.75\baselineskip}
In this section, we evaluate our \alphadpp sampler on a
benchmark\footnote{\url{https://github.com/LCSL/dpp-vfx}}
introduced by \cite{derezinski2019exact} (see
\Cref{a:experiments}). The benchmark uses
subsets of the infinite MNIST dataset \cite{loosli-canu-bottou-2006}
with $d = 784$ and $n$ varying up to $10^6$.
All experiments are executed on a 28 core Xeon E5-2680 v4. Each experiment is repeated multiple times, and we report mean
values and a 95\% confidence interval.

\textbf{Baselines:} we compare \alphadpp with \dppvfx \cite{derezinski2019exact}, an MCMC sampler \cite{rayleigh-mcmc}
and a sampler based on eigendecompositions \cite{dpp-independence, gillenwater2014approximate}.
All algorithms  are implemented in \texttt{python} as part
\ifisarxiv
\else
\footnote{We provide our modified version of the library which includes \alphadpp in the supplementary material.}
\fi
of \texttt{DPPy} \cite{gautier2018dppy}.
Due to their similar input, we use the same oversampling parameters (see \Cref{a:experiments}) for \alphadpp and \dppvfx.
We run the MCMC sampler for $\cO(nk)$ iterations to guarantee mixing \cite{rayleigh-mcmc}. For more details
on hyperparameter tuning we refer to \Cref{a:experiments}.

\textbf{Results} \ We begin by reporting results on a smaller subset of data (\Cref{fig:small-scale})
where even the non-efficient samplers can be run. We use an \texttt{rbf}
similarity with $\sigma = \sqrt{3d}$, and set $k=10$ to match the number of
digit classes in MNIST. Note that for $n = 70000$ \pbless estimates
$\deff(\L) \approx 300$, validating our assumption of $\deff(\L) \gg k$.
Thanks to this mismatch, we can see how \alphadpp maintains a \emph{constant}
runtime as $n$ grows, and increasingly matches or outpaces competing baselines as $n$ grows.
In particular, it becomes faster than the eigendecomposition based sampler (which cannot scale beyond $n = 24000$)
or the MCMC sampler. However, the gap is still sufficiently small that
\dppvfx, the previously fastest $k$-DPP sampler
available, remains competitive.

For larger datasets we consider only the scalable samplers, \alphadpp and
\dppvfx. We consider again an \texttt{rbf} similarity, but this time we choose $n$ up
to $10^6$ and $\sigma = \sqrt{10}$. This further increases the gap between $k$ and $\deff(\L)$,
with \pbless estimating $\deff(\L) \approx 1000$.
We report results in \Cref{fig:large-scale}, with runtime shown in
log-scale. In this regime, the gap between \dppvfx and
\alphadpp widens, as \dppvfx
cannot use rescaling to reduce the final dictionary size from $\deff(\L)$ to
$\deff(\hat\alpha\L) \approx k$, and has to compute $n$ marginal probabilities
since it does not leverage uniform intermediate subsampling.
In particular, thanks to the uniform sampling step, we see that \alphadpp's
runtime does not grow as $n$ grows, since all the expensive computations
are performed in the small intermediate subset which is hardly
sensitive to $n$. We note that, due to using a smaller dictionary,
\alphadpp requires about 2-5x more trials in the rejection sampling
step, which leads to larger variance in the runtime.

\begin{wrapfigure}[12]{R}{0.43\textwidth}
  \begin{minipage}{\linewidth}
    \vspace{-5mm}
\centering
\includegraphics[width=.95\textwidth]{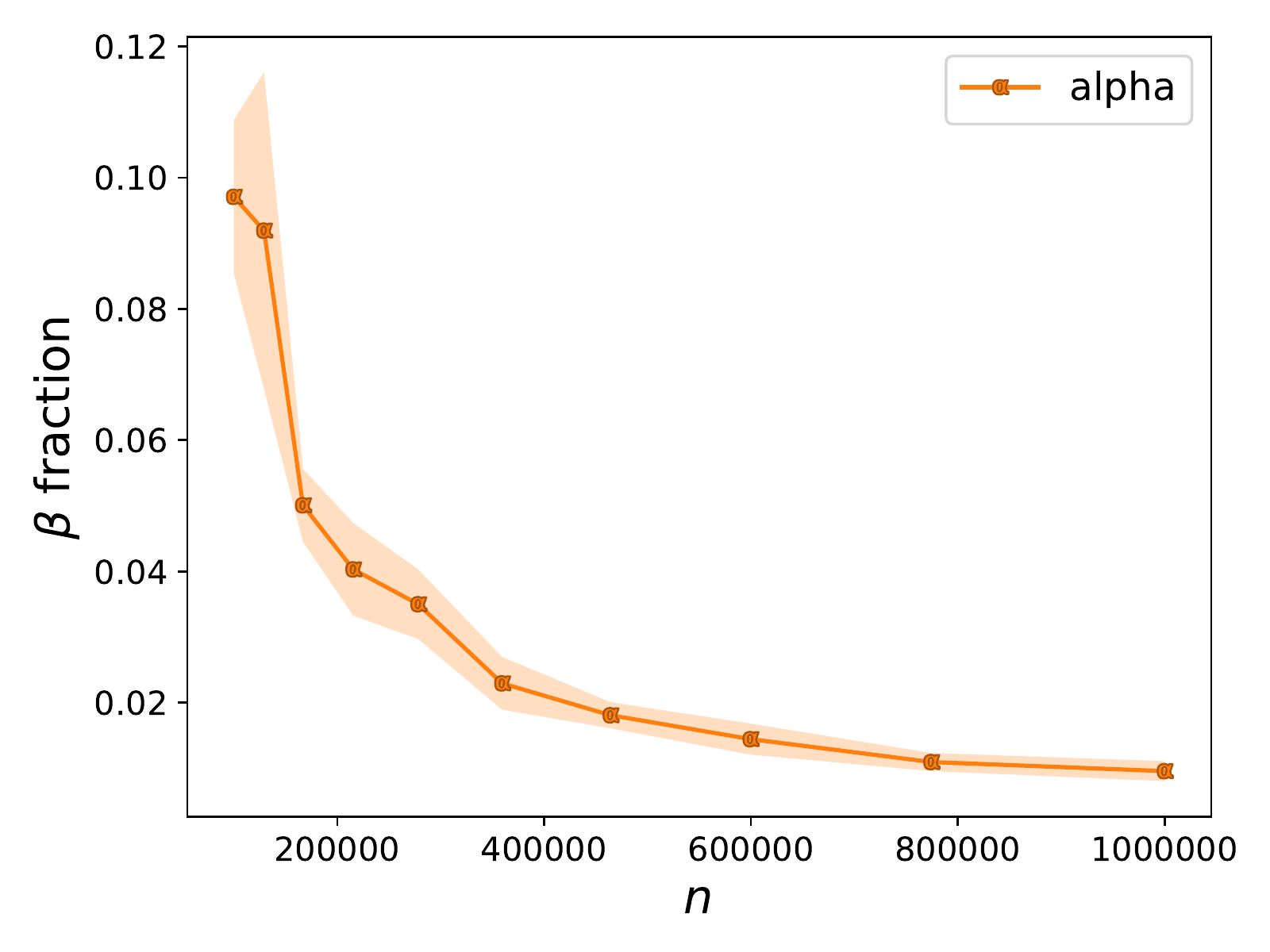}
\vspace{-2mm}
\captionsetup{font=small}
\captionof{figure}{Fraction of items observed by \alphadpp.}\label{fig:rejections}
\vspace{-4mm}
    \end{minipage}
    \end{wrapfigure}

In \Cref{fig:rejections}, we report the fraction of data that
is observed by \alphadpp in the large scale experiment. This
quantity, denoted as $\beta$ in \Cref{thm:informal-main}, is
responsible for much of the computational gains of the algorithm over
\dppvfx, reported in \Cref{fig:large-scale}. Note that the remaining
$1-\beta$ portion of the data does not ever need to be loaded into
the program's memory, which leads to a significant reduction in
memory accesses. We observe that as the
data size increases, the fraction of items observed by \alphadpp goes down to as
little as $1$\% for $n=10^6$, which is why the
runtime of \alphadpp stays roughly flat, whereas the runtime of
$\dppvfx$ grows.

\newpage
\section*{Broader impact}
DPPs were discovered in the 70s by Odile Macchi to model repulsion of
particle distributions in fermions, so improvements in samplers may help
in modelling physical simulations. In bringing faster DPP samplers to
machine learning we aim to enable a
better handling of diversity through this rigorous theoretical framework.
\begin{ack}
  MD thanks the NSF for funding via the NSF TRIPODS program.
\end{ack}

\bibliographystyle{plainnat}

\newpage
\appendix
\section{Correctness and efficiency of \alphadpp (\Cref{alg:alpha-dpp-sampler})}\label{a:a-dpp-proofs}
In this section we prove the theorems stated in \Cref{s:alpha-sampler,s:binary-search} claiming the correctness and efficiency of \dppvfx.
In particular, we split \Cref{thm:alpha-sampling} into two parts.

\begin{lemma}\label{lem:a-dpp-exact}
Given any psd matrix $\L$, dictionary $\coldict$, positive weights $\W$, $r \geq 1,$ and positive $\alpha > 0$, \alphadpp returns
$S \sim \DPP(\alpha\L)$.
\end{lemma}

\begin{lemma}\label{lem:a-dpp-efficient}
If $r \geq \deff(\alpha\L) \geq 1/2$, $\coldict$ and $\W$ are $(1/\deff(\alpha\L), \alpha)$-accurate,
$\coldict$ satisfies $|\coldict| \leq 10\deff(\alpha\L)$, and
$\deff(\alpha\Lbh) \leq 10\deff(\alpha\L)$, then with probability $1 -
\delta$, \alphadpp (\Cref{alg:alpha-dpp-sampler}) runs in
time\vspace{-1mm}
\begin{align*}
  \cO\Big(\left[ \min\{\alpha\kappa^2\deff(\alpha\L), 1\}\cdot n \cdot
  \deff(\alpha\L)^6\log^2(n/\delta) +
  \deff(\alpha\L)^9\log^3(n/\delta)\right ] \cdot \log(1/\delta)\Big).
  \end{align*}
\end{lemma}

\subsection{Notation}
We start by introducing some additional notation. First, let us describe the so-called kernel-based view of DPPs.
We associate with our similarity matrix $\L$ a similarity function or a
kernel\footnote{Note that we are defining the kernel $\kerfunc$ as a function on indices, but since we focus on DPPs defined on PSD matrices, everything can be immediately extended to any standard PSD kernel $\kerfunc(\cdot, \cdot): \statespace\times\statespace\rightarrow\R$ defined on an arbitrary input space $\statespace$.}
function
$\kerfunc(\cdot,\cdot):[n]\times[n] \rightarrow \R$ such that
$\kerfunc(i,j)$ is equal to the $(i,j)$-th entry of $\L$.

We also generalize the notation just defined in a way that given multi-sets $A$ and $B$,
$\kerfunc(A, B) \defeq \L_{A,B}$ returns the matrix
containing the corresponding rows and columns of $\L$. Note that if $A$ or $B$
contains duplicates (\eg the $i$-th index appears twice in $A$) the matrix
$\kerfunc(A ,B)$ will consequently contain duplicate rows and columns. Finally,
note that in this notation the original matrix can be written as $\L = \kerfunc([n], [n])$.

We also denote with $\featmap(\cdot): [n] \rightarrow \R^D$
the so-called feature map associated with $\L$ and $\kerfunc(\cdot, \cdot)$ such that
$\kerfunc(i, j) = \featmap(i)^\transp\featmap(j)$, where $D$ can be
arbitrarily large or
infinite.\footnote{Again we focus on a feature map from indices to a finite dimensional space. All the results can be immediately extended to a feature map $\featmap(\cdot): \statespace \rightarrow \rkhs$ that maps from an arbitrary input space into a reproducing kernel Hilbert space, \eg Gaussian kernel and Gaussian feature maps.
Notice that in our setting given a PSD matrix $\L$ the eigenspace of $\L$ suffices to construct an appropriate feature map $\featmap(\cdot): [n] \rightarrow \R^n$ with $D = n$. In particular, we have an explicit expression for $\featmap(\cdot)$ based on the eigendecomposition
$\L = \U\Sigma\U^\transp$ of $\L$. Since~$\L$ is psd, $\Sigma$ is a diagonal
matrix with non-negative entries, and we can define $\Sigma^{+/2}$ as the
square root of its pseudo-inverse. Then the feature map becomes
$\featmap(\cdot) \defeq \Sigma^{+/2}\U^\transp\kerfunc([n], \cdot)$.
A similar argument can be made using the Cholesky decomposition of $\L$.
}
Just as with $\kerfunc$, we also extend $\featmap(\cdot)$ to operate on
multi-set, such that given $A = \{i_1, \ldots, i_m\}$ (potentially with duplicates $i_j = i_l$),
we have $\featmap(A) = [\featmap(i_1), \ldots, \featmap(i_m)]^\transp \in \R^{m \times D}$.

Using the above notation, we have $\L = \kerfunc([n],[n]) = \featmap([n])\featmap([n])^\transp$. Note also that the corresponding operator $\featmap([n])^\transp\featmap([n])$ can be decomposed as a sum of outer products $\featmap([n])^\transp\featmap([n]) = \sum_{i=1}^n\featmap(i)\featmap(i)^\transp$.

We also use the following notation to indicate common sampling distributions:
\begin{itemize}[leftmargin=*]
\vspace{-0.2cm}
\item $u \sim \Poisson(\lambda)$ as a non-negative integer sampled from a Poisson distribution with intensity $0<\lambda$;
\item $\unifset \sim \Uniform(u, [n])$ as a set of size $u$ sampled uniformly i.i.d.\,with replacement from $[n]$;
\ie $\unifset = (\unifset_1, \ldots, \unifset_u) \simiid (1/n, \ldots, 1/n)$.
\item $z \sim \Bernoulli(p)$ as the $\{0,1\}$ r.v.\,sampled from a Bernoulli distribution w.p.~$0 \leq p \leq 1$;
\item $s \sim \Binomial(k,p)$ as the non-negative integer in the range $[0,k]$ sampled from a Binomial distribution with $0 \leq k$ Bernoulli repetitions each with probability $0 \leq p \leq 1$
\item $(s_1,\ldots, s_n) \sim \Multinoulli(k, \{p_i\}_{i=1}^n)$ as the vector of positive integers $[0,k]^n$ with $0 \leq k$ sampled according to $\P\left((s_1, \ldots, s_n)\right) = \frac{n!}{\prod_{i=1}^n s_i!} \prod_{i=1}^n p_i^{s_i}$  such that $\sum_{i=1}^n s_i = k$.
\item $\sigma \sim \Multinomial(k, \{p_i\}_{i=1}^n)$ as a set of size $k$ sampled i.i.d.\,with replacement from $[n]$ according to probabilities
$0 \leq p_i \leq 1$ with $\sum_{i=1}^n p_i = 1$, \ie $\sigma = (\sigma_1, \ldots, \sigma_k) \simiid (p_1, \ldots, p_n)$.
\end{itemize}

\subsection{Proof of \Cref{lem:a-dpp-exact} (exact sampling)}
To prove that \alphadpp is an exact sampler we show that $\sigma$ is sampled
according to an appropriate $\Vol$, and that therefore we can invoke
\Cref{t:composition}.

\begin{proof}[Proof of \Cref{lem:a-dpp-exact}]
Given the approximate marginals $l_i$ from \Cref{eq:rls-approx}, let us denote with $\adeff(\alpha\L) \defeq \sum_{i=1}^n l_i$
their sum, or approximate effective dimension.
Note that \Cref{alg:alpha-dpp-sampler}
\emph{never}  computes $\adeff(\alpha\L)$ explicitly, nor does it
compute all approximate marginals $l_i$.
Nonetheless, our first claim is that the inner loop of
\alphadpp is proposing a candidate $\sigma$ sampled according
to the approximate marginals $l_i$ even without computing them all.
\begin{lemma}\label{lem:equivalent-sampling}
The set $\sigma$ generated by \Cref{alg:alpha-dpp-sampler} before \Cref{line:acc} is distributed as
\begin{align*}
\sigma = \{\sigma_1, \dots, \sigma_t\} \simiid (l_1/\adeff(\alpha\L), \dots , l_n/\adeff(\alpha\L)), \quad\quad t \sim \Poisson\left(re^{1/r}\adeff\left(\alpha\L\right)\right).
\end{align*}
\end{lemma}
Then, we show that
the rejection sampling step of \Cref{line:acc} is valid.
\begin{lemma}\label{lem:rejection-alpha-valid}
Given any psd matrix $\L$, dictionary $\coldict$, positive weights $\W$, $r \geq 1,$ and positive $\alpha > 0$,
the acceptance
probability $\tfrac{\ee^{\st}\det(\I+ \alpha\Lbt_\sigma)}{\ee^{t/r}\det(\I+\alpha\Lbh)} \leq 1$ is valid.
\end{lemma}
Let $\sigmat$ denote the random variable distributed as
$\sigma$ is after exiting the repeat loop.
Combining \Cref{lem:rejection-alpha-valid} with the fact that $t \sim \Poisson(re^{1/r}\adeff(\alpha\L))$ is a Poisson r.v.\,it
follows that
\begin{align*}
\Pr(\sigmat\in A)
&\propto \E_{\sigma}
\bigg[\one_{[\sigma\in A]}\frac{\ee^{\st}\det(\I+\Lbt_\sigma)}{\ee^{t/r}\det(\I+\Lbh)}\bigg]\\
&\propto \sum_{t=0}^\infty
\frac{(r\,\ee^{1/r}\adeff(\alpha\L))^t}{\ee^{r\,\ee^{1/r}\adeff(\alpha\L)}\,t!}\cdot
     \ee^{-t/r}\,
     \E_\sigma\big[\one_{[\sigma\in A]}\det(\I+\Lbt_\sigma)\mid t\big]
\\ &\propto \E_{t'}\Big[\E_\sigma\big[\one_{[\sigma\in A]}\det(\I+\Lbt_\sigma)\mid
     t=t'\big]\Big]\quad\text{for } t'\sim\mathrm{Poisson}(r\adeff(\alpha\L)),
\end{align*}
which matches the numerator of \Cref{d:r-dpp} for a $\Vol^{r\adeff(\alpha\L)}_{\{l_i/\adeff(\alpha\L)\}_{i=1}^n}$.
All that remains is to show that the distribution integrates properly, \ie the denominator also matches.
We do this by generalizing a determinantal equality
to our modified reweighting.
\begin{proposition}[\cite{dpp-intermediate}]\label{prop:expected-det-poisson}
If $t \sim \Poisson(r\adeff(\alpha\L))$ and
$(\sigma_1, \dots, \sigma_t) \simiid \left(\tfrac{l_1}{\adeff(\alpha\L)},\dots, \tfrac{l_n}{\adeff(\alpha\L)}\right)$ then
\begin{align*}
\textstyle
\E_{t,\sigma}\left[\det\left(\I + \wt{\Lb}_{\sigma}\right)\right]
= \det\left(\I + \Lb\right).
\end{align*}
\end{proposition}
This shows that $\sigmat\sim
\Vol^{r\adeff(\alpha\L)}_{\{l_i/\adeff(\alpha\L)\}_{i=1}^n}$. The claim follows from \Cref{t:composition}.
\end{proof}
\begin{proof}[Proof of \Cref{lem:equivalent-sampling}]
Before starting, we will use two well known connections of the Poisson distribution
with GenBinomial and Binomial r.v.\,\cite{katti1968handbook}.
The first useful Poisson property is that for any set of positive weights $\{\lambda_i\}_{i=1}^n$ the random variables
$X_1 \sim \Poisson(\lambda_1), \dots, X_n \sim \Poisson(\lambda_n)$
and the random variables
\begin{align*}
\textstyle
k \sim \Poisson\big(\sum_{i=1}^n\lambda_i\big),  \quad\quad  \{X_1, \dots, X_n\}|k \sim \Multinoulli\left(k, \tfrac{\lambda_i}{\sum_{i=1}^n\lambda_i}\right)
\end{align*}
are equally distributed.
Note that for this identity
to hold we do not need to explicitly compute $\sum_{i=1}^n\lambda_i$, as we can simply sample $n$ Poisson r.v.-s and obtain the normalization effect in the GenBinomial sample for free using the conditioning on $k$. The second useful Poisson property we will use is that if
$k \sim \Poisson(\lambda)$ and $Y|k \sim \Binomial(k, p)$ then $Y \sim \Poisson(\lambda \cdot p)$.

Let us denote with $u_i = \sum_{j=1}^u \indfunc\{\unifset_j = i\}$ the multiplicity
of index $i$ in $\unifset$, such that we have a set of~$n$ random variables
$\{u_i\}_{i=1}^n$ and that $u = \sum_{i=1}^n u_i$.
Then, from the previous relationship, we can instantly see that
sampling $u_i \sim \Poisson(re^{1/r}b)$ is equivalent to sampling
\begin{align*}
&u \sim \Poisson\big(\sum_{i=1}^n re^{1/r}\alpha\kappa^2\big) = \Poisson(re^{1/r}n\alpha\kappa^2),\\
&\{u_1, \dots, u_n\}|u \sim \Multinoulli\left(u, \tfrac{re^{1/r}\alpha\kappa^2}{re^{1/r}n\alpha\kappa^2}\right) = \Multinoulli\left(u, \left\{\tfrac{1}{n}\right\}\right)\cdot
\end{align*}
We can now connect $u_i$ and $\unifset$. In particular
sampling $\unifset|u \sim \Uniform(u,[n])$ is equivalent
to sampling $\{u_1, \dots, u_n\}|u \simiid \Multinoulli\left(u, \left\{\tfrac{1}{n}\right\}\right)$
and then adding $u_i$ copies of $i$ to $\unifset|u$ for each $i \in [n]$.

Starting from this characterization, let us now denote with
$s_i = \sum_{j=1}^t \indfunc\{\sigma_j = i\}$
the multiplicity of index $i$ in $\sigma$, such that we have a set of $n$ random variables
$\{s_i\}_{i=1}^n$ and that $t = \sum_{i=1}^n s_i$.
We can now formally describe \Cref{line:reject-binomial} of \Cref{alg:alpha-dpp-sampler} as a binomial sampling step:
first we sample $u_i \sim \Poisson(re^{1/r}\alpha\kappa^2)$, and then we sample $s_i|u_i \sim \Binomial(u_i, l_i/(\alpha\kappa^2))$.
To see this, we can just sum over all $z_j$ that correspond to the $i$-th
element, of which we have exactly $u_i$, and remember that a sum of i.i.d. Bernoulli
is a Binomial. We also have to take care of the fact that the Binomial probability
is well defined, \ie smaller than 1, but it is easy to see that
$l_i \leq \alpha[\L]_{i,i} \leq \alpha\kappa^2$ and $l_i/(\alpha\kappa^2) \leq 1$.
We can now use the second fact about Poissons, namely that sampling $u_i \sim \Poisson(re^{1/r}b)$
and then $s_i|u_i \sim \Binomial(u_i, \tfrac{l_i}{b})$ is equivalent to sampling $s_i \sim \Poisson(re^{1/r}b\cdot\tfrac{l_i}{b}) = \Poisson(re^{1/r}l_i)$.

Finally we can once again use the equivalence between Poisson and GenBinomial sampling to see
that sampling $s_i \sim \Poisson(re^{1/r}l_i)$ for each $i \in [n]$ is equivalent to sampling
\begin{align*}
\textstyle
t \sim \Poisson\big(\sum_{i=1}^n ql_i\big) = \Poisson(re^{1/r}\adeff(\alpha\L)),\quad
\{s_1, \dots, s_n\}|t \sim \Multinoulli\left(t, \frac{l_i}{\adeff(\alpha\L)}\right)\CommaBin
\end{align*}
which in turn implies that by adding $s_i$ copies of the index $i$ to $\sigma$, which is what \Cref{alg:alpha-dpp-sampler} is doing,
we are sampling according to
\begin{align*}
\sigma = \{\sigma_1, \dots, \sigma_t\} \simiid (l_1/\adeff(\alpha\L), \dots , l_n/\adeff(\alpha\L)), \quad\quad t \sim \Poisson(re^{1/r}\adeff(\alpha\L)),
\end{align*}
without ever explicitly computing $\adeff(\alpha\L)$.
 \begin{algorithm}[t]
   \caption{\alphadpp reformulation in terms of $u_i$ and $s_i$}
 \label{alg:alpha-dpp-sampler-reject}
 \begin{algorithmic}[1]
 \renewcommand{\algorithmicrequire}{\textbf{Input:}}
 \REQUIRE $\alpha$, $\L$, an $m$-element dictionary $\coldict$, weight matrix $\W \in \R^{m \times m}$,
 \ $r \geq 1$
\STATE Set $\Lbh = \W^{1/2}\L_{\coldict,\coldict}\W^{1/2} \in \R^{m \times m}$
\REPEAT
\FOR{$i = \{1, \dots, n\}$}
\STATE Sample $u_i \sim \Poisson(re^{1/r}\alpha\kappa^2)$
\IF{$u_i > 0$}
\STATE Compute $l_{i}$ using \Cref{eq:rls-approx}
\STATE Sample $s_i \sim \Binomial(u_i, l_{i}/(\alpha\kappa^2))$
\ELSE
\STATE Set $s_i = 0$
\ENDIF
\STATE Add $s_i$ copies of $i$ to $\sigma$
\ENDFOR
\STATE Set $t = |\sigma|$, $[\Lbt_{\sigma}]_{ij} = \tfrac{1}{r\sqrt{l_{\sigma_i} l_{\sigma_j}}}[\L]_{\sigma_i\sigma_j}$
\STATE Sample $\textit{Acc}\sim\!
   \Bernoulli\Big(\ee^{\st - t/r}\det(\I+ \alpha\Lbt_\sigma)/{\det(\I+\alpha\Lbh)}
 \Big)$
 \UNTIL{$\textit{Acc}=\text{true}$ \STATE Sample $\St\sim \DPP\big(\alpha\Lbt_\sigma\big)$}
   \RETURN $S = \{\sigma_i: i\!\in\! \St\}$
 \end{algorithmic}
 \end{algorithm}

 \begin{algorithm}[t]
   \caption{\alphadpp reformulation in terms of $s_i$}
 \label{alg:alpha-dpp-sampler-reject-slow}
 \begin{algorithmic}[1]
 \renewcommand{\algorithmicrequire}{\textbf{Input:}}
 \REQUIRE $\alpha$, $\L$, an $m$-element dictionary $\coldict$, weight matrix $\W \in \R^{m \times m}$,
 \ $r \geq 1$
\REPEAT
\FOR{$i = \{1, \dots, n\}$}
\STATE Compute $l_{i}$ using \Cref{eq:rls-approx}
\STATE Sample $s_i \sim \Poisson(re^{1/r}l_{i})$
\STATE Add $s_i$ copies of $i$ to $\sigma$
\ENDFOR
\STATE Set $t = |\sigma|$, $[\Lbt_{\sigma}]_{ij} = \tfrac{1}{r\sqrt{l_{\sigma_i} l_{\sigma_j}}}[\L]_{\sigma_i\sigma_j}$
\STATE Sample $\textit{Acc}\sim\!
   \Bernoulli\Big(\ee^{\st - t/r}\det(\I+ \alpha\Lbt_\sigma)/{\det(\I+\alpha\Lbh)}
 \Big)$
 \UNTIL{$\textit{Acc}=\text{true}$ \STATE Sample $\St\sim \DPP\big(\alpha\Lbt_\sigma\big)$}
   \RETURN $S = \{\sigma_i: i\!\in\! \St\}$
 \end{algorithmic}
 \end{algorithm}

For completeness, we also include the two implicit reformulations of
\Cref{alg:alpha-dpp-sampler} that we just described as \Cref{alg:alpha-dpp-sampler-reject}
and \Cref{alg:alpha-dpp-sampler-reject-slow}. Note that all three algorithms
are strictly equivalent, but depending on the actual implementation they have
different complexities. For example, \Cref{alg:alpha-dpp-sampler-reject-slow}
needs to compute all marginals in advance. We chose to include \Cref{alg:alpha-dpp-sampler}
in the main paper as the version that more clearly highlights the uniform
sampling step.
\end{proof}

\begin{proof}[Proof of \Cref{lem:rejection-alpha-valid}]
The first reason we introduced the kernel-based DPP notation is to be able to
succinctly use Sylvester's identity to equate determinants in the
matrix and feature view of the DPP, \ie
\begin{align*}
\textstyle
\det\left(\I + \L\right) = \det\left(\I + \featmap([n])\featmap([n])^\transp\right)
= \det\left(\I + \featmap([n])^\transp\featmap([n])\right)
= \det\left(\I + \sum_{i=1}^n\featmap(i)^\transp\featmap(i)\right),
\end{align*}
where the size of the identity matrix\footnote{Or an identity operator on an RKHS in general} $\I$ is either $n$ or $D$ and it is clear from the context.
Similarly, the denominator $\det(\I + \alpha \Lbh) = \det(\I + \alpha \W^{1/2}\L_{\coldict,\coldict}\W^{1/2})$ in the rejection loop becomes
\begin{align*}
\det(\I + \alpha \W^{1/2}\L_{\coldict,\coldict}\W^{1/2})
=\det(\I + \alpha \W^{1/2}\featmap(\coldict)\featmap(\coldict)^\transp\W^{1/2})
=\det(\I + \alpha \featmap(\coldict)^\transp\W\featmap(\coldict)).
\end{align*}
Finally, given $\sigma$ let us denote with $\wt{\featmap}(i) = \tfrac{1}{\sqrt{rl_{i}}}\featmap(i)$ a rescaled feature map, where once again we extend
$\wt{\featmap}(\sigma) =
\text{Diag}(\tfrac{1}{\sqrt{rl_{\sigma_i}}})_{i=1}^m\featmap(\sigma)$
to multi-sets. Then the numerator in the rejection loop becomes
\begin{align*}
\det(\I + \alpha \Lbt_{\sigma})
& =\det(\I + \alpha \wt{\featmap}(\sigma)\wt{\featmap}(\sigma)^\transp)\\
&=\det(\I + \alpha \wt{\featmap}(\sigma)^\transp\wt{\featmap}(\sigma))
=\det\left(\I + \alpha \sum\nolimits_{j=1}^t\tfrac{1}{rl_{\sigma_j}}\featmap(\sigma_j)\featmap(\sigma_j)^\transp\right).
\end{align*}
The second reason we introduce this notation is that the formulation of the approximate marginals $l_i$
is much simplified and becomes (see \cite{calandriello_disqueak_2017,NIPS2018_7810} for details)
\begin{align}\label{eq:rls-reformulation}
l_i
= \alpha[\L - \alpha\L_{[n],\coldict}^\transp(\alpha\L_{\coldict,\coldict} + \W^{-1})^{-1}\L_{[n],\coldict}]_{i,i}
= \alpha\featmap(i)^\transp(\I + \alpha\featmap(\coldict)^\transp\W\featmap(\coldict))^{-1}\featmap(i).
\end{align}

Using the kernel-based
view of DPPs and the reformulation of most quantities, we can now move from characterizing the distribution of $\sigma$,
to computing the final acceptance probability $\P(\textit{Acc}|\sigma)$.
In particular, to guarantee correctness we must guarantee that the rejection
step is valid, \ie that the acceptance probability is bounded by 1.
For this we rewrite the acceptance condition as
\begin{align*}
\frac{\det(\I + \alpha\Lbt_{\sigma})}{\det(\I + \alpha\Lbh)}
= \frac{\det(\I + \alpha \wt{\featmap}(\sigma)^\transp\wt{\featmap}(\sigma))}{\det(\I + \alpha \featmap(\coldict)^\transp\W\featmap(\coldict))}\cdot
\end{align*}

Similarly to \cite{derezinski2019exact}, we can use the inequality
$\det(\I + \A) \leq \exp\{ \tr(\A) \}$, which follows immediately by applying
the bound $1 + x \leq e^x$ to each singular value of $\A$. We obtain
\begin{align*}
&\frac{\det(\I + \alpha \wt{\featmap}(\sigma)^\transp\wt{\featmap}(\sigma))}{\det(\I + \alpha \featmap(\coldict)^\transp\W\featmap(\coldict))}\\
&=
\det\left((\I + \alpha \wt{\featmap}(\sigma)^\transp\wt{\featmap}(\sigma))(\I + \alpha \featmap(\coldict)^\transp\W\featmap(\coldict))^{-1}\right)
\\
& =
\det\left(\I + (\alpha \wt{\featmap}(\sigma)^\transp\wt{\featmap}(\sigma) - \alpha \featmap(\coldict)^\transp\W\featmap(\coldict))(\I + \alpha \featmap(\coldict)^\transp\W\featmap(\coldict))^{-1}\right)\\
&\leq \exp\left\{\tr\left(
(\alpha \wt{\featmap}(\sigma)^\transp\wt{\featmap}(\sigma) - \alpha \featmap(\coldict)^\transp\W\featmap(\coldict))(\I + \alpha \featmap(\coldict)^\transp\W\featmap(\coldict))^{-1}
\right)\right\}\\
&= \exp\{\underbracket{\tr\left(\alpha\wt{\featmap}(\sigma)^\transp\wt{\featmap}(\sigma)(\I + \alpha \featmap(\coldict)^\transp\W\featmap(\coldict))^{-1}\right)}_{(a)} - \underbracket{\tr\left(\alpha \featmap(\coldict)^\transp\W\featmap(\coldict)(\I + \alpha \featmap(\coldict)^\transp\W\featmap(\coldict))^{-1}\right)}_{(b)}\}.
\end{align*}
For $(b)$, we can see that by definition $\tr\left(\alpha \featmap(\coldict)^\transp\W\featmap(\coldict)(\I + \alpha \featmap(\coldict)^\transp\W\featmap(\coldict))^{-1}\right) = \st$. For $(a)$, we have
\begin{align*}
(a) &= \tr\left(\alpha\wt{\featmap}(\sigma)^\transp\wt{\featmap}(\sigma)(\I + \alpha \featmap(\coldict)^\transp\W\featmap(\coldict))^{-1}\right)
= \tr\left(\alpha\wt{\featmap}(\sigma)(\I + \alpha \featmap(\coldict)^\transp\W\featmap(\coldict))^{-1}\wt{\featmap}(\sigma)^\transp\right)\\
&= \sum_{j=1}^t\alpha\wt{\featmap}(\sigma_j)^\transp(\I + \alpha \featmap(\coldict)^\transp\W\featmap(\coldict))^{-1}\wt{\featmap}(\sigma_j)
= \sum_{j=1}^t\frac{1}{rl_{\sigma_j}}\alpha\featmap(\sigma_j)^\transp(\I + \alpha \featmap(\coldict)^\transp\W\featmap(\coldict))^{-1}\featmap(\sigma_j)\\
&= \sum_{j=1}^t\frac{1}{rl_{ \sigma_j }}l_{\sigma_j} = \frac{t}{r}.
\end{align*}
Putting $(a)$ and $(b)$ together we have
\begin{align*}
\P(\textit{Acc} = \text{true}|\sigma) =
&\frac{\det(\I + \alpha \wt{\featmap}(\sigma)^\transp\wt{\featmap}(\sigma))}{\det(\I + \alpha \featmap(\coldict)^\transp\W\featmap(\coldict))}\cdot{\exp\left\{\st - \frac{t}{r}\right\}}\\
&\leq {\exp\left\{\frac{t}{r} - \st\right\}}\cdot{\exp\left\{\st - \frac{t}{r}\right\}}
= e^0 = 1.
\end{align*}
\end{proof}

\begin{proof}[Proof of \Cref{prop:expected-det-poisson}]
We first rewrite the equality as
\begin{align*}
\E_{t,\sigma}\bigg[\det\Big(\I + \sum_{j=1}^t\frac{1}{rl_{\sigma_j}}\featmap(\sigma_j)\featmap(\sigma_j)^\transp\Big)\bigg]
= \det\Big(\I + \sum_{i=1}^n\featmap(i)\featmap(i)^\transp\Big)\cdot
\end{align*}
\citet{dpp-intermediate} showed the following identity when sampling $t \sim \Poisson(r)$ and then sampling a multi-set $\sigma$ with $t$ elements i.i.d.\,from any arbitrary distribution,
\begin{align*}
\E_{t,\sigma}\left[\det\left(\I + \featmap(\sigma)^\transp\featmap(\sigma)\right)\right]
= \det\left(\I + r\E_{\sigma_1}\left[\featmap(\sigma_1)\featmap(\sigma_1)^\transp\right]\right).
\end{align*}
Applying this to our $t\sim \Poisson(r\adeff(\alpha\L))$ and the distribution of $\sigma$ we have
\begin{align*}
\E_{t,\sigma}\bigg[\det\Big(\I + \sum_{j=1}^t\frac{1}{rl_{\sigma_j}}\featmap(\sigma_j)\featmap(\sigma_j)^\transp\Big)\bigg]
&=\det\left(\I + r\adeff(\alpha\L)\E_{\sigma_1}\left[\featmap(\sigma_1)\featmap(\sigma_1)^\transp\right]\right)\\
&= \det\Big(\I + r\adeff(\alpha\L)\sum_{i=1}^n\frac{l_i}{\adeff(\alpha\L)}\frac{1}{rl_{i}}\featmap(\sigma_i)\featmap(\sigma_i)^\transp\Big)\\
&= \det\Big(\I + \sum_{i=1}^n\featmap(i)\featmap(i)^\transp\Big).
\end{align*}
\end{proof}
\subsection{Proof of \Cref{lem:a-dpp-efficient} (efficiency)}
\begin{proof}[Proof of \Cref{lem:a-dpp-efficient}]
We need to lower bound the acceptance probability
$\P(\textit{Acc}|\sigma)$. Note that
this is equivalent to lower bounding
$\E[\textit{Acc} = \text{true}]$ since it is a $\{0,1\}$ random variable.

\begin{align*}
\P(\textit{Acc} = \text{true})
&= \E_{\sigma,t}\left[
\frac{e^{\st}\det(\I + \alpha\wt{\featmap}(\sigma)^\transp\wt{\featmap}(\sigma))}{e^{t/r}\det(\I + \alpha\Lbh)}\right]
= \frac{e^{\st}}{\det(\I + \alpha\Lbh)}\E_{\sigma,t}\left[
\frac{\det(\I + \alpha\wt{\featmap}(\sigma)^\transp\wt{\featmap}(\sigma))}{e^{t/r}}\right]\\
&= \frac{e^{\st}}{\det(\I + \alpha\Lbh)}\cdot\sum_{t=0}^{\infty}
\E_{\sigma}\big[\det(\I + \alpha\wt{\featmap}(\sigma)^\transp\wt{\featmap}(\sigma))|\; t\;\big]
\tfrac{1}{e^{t/r}}\tfrac{\left(re^{1/r}\adeff(\alpha\L)\right)^t}{t!\cdot e^{re^{1/r}\adeff(\alpha\L)}},
\end{align*}
where we expanded the expectation with respect to $t \sim
\Poisson(r e^{1/r}\adeff(\alpha\L))$.
Focusing on the last term we have
\begin{align*}
\frac{1}{e^{t/r}}\frac{\left(re^{1/r}\adeff(\alpha\L)\right)^t}{t!\cdot e^{re^{1/r}\adeff(\alpha\L)}}.
= \frac{1}{e^{t/r}}\frac{r^te^{t/r}\adeff(\alpha\L)^t}{t!\cdot e^{re^{1/r}\adeff(\alpha\L)}}
= \frac{r^t\adeff(\alpha\L)^t}{t!\cdot e^{re^{1/r}\adeff(\alpha\L)}}
= \frac{(r\adeff(\alpha\L))^t}{t!\cdot e^{r\adeff(\alpha\L)}}\frac{e^{r\adeff(\alpha\L)}}{e^{re^{1/r}\adeff(\alpha\L)}}.
\end{align*}
Recognizing that $\tfrac{(r\adeff(\alpha\L))^t}{t!\cdot e^{r\adeff(\alpha\L)}}$ is the density
of a $t \sim \Poisson(r\adeff(\alpha\L))$, we can apply \Cref{prop:expected-det-poisson},
\begin{align*}
\P(\textit{Acc} = \text{true})
&= \frac{e^{\st}}{\det(\I + \alpha\Lbh)}\cdot\sum_{t=0}^{\infty}
\E_{\sigma}\big[\det(\I + \alpha\wt{\featmap}(\sigma)^\transp\wt{\featmap}(\sigma))|\; t\;\big]
\tfrac{1}{e^{t/r}}\tfrac{\left(re^{1/r}\adeff(\alpha\L)\right)^t}{t!\cdot e^{re^{1/r}\adeff(\alpha\L)}}\\
&= \frac{e^{\st}}{\det(\I + \alpha\Lbh)}\frac{e^{r\adeff(\alpha\L)}}{e^{re^{1/r}\adeff(\alpha\L)}}\cdot\sum_{t=0}^{\infty}
\E_{\sigma}\big[\det(\I + \alpha\wt{\featmap}(\sigma)^\transp\wt{\featmap}(\sigma))|\; t\;\big]
\frac{(r\adeff(\alpha\L))^t}{t!\cdot e^{r\adeff(\alpha\L)}}\\
&= \frac{e^{\st}}{\det(\I + \alpha\Lbh)}\frac{e^{r\adeff(\alpha\L)}}{e^{re^{1/r}\adeff(\alpha\L)}}\cdot\det(\I + \alpha{\featmap}([n])^\transp{\featmap}([n]))\\
&= \frac{{e^{\st}}e^{r\adeff(\alpha\L)}}{e^{re^{1/r}\adeff(\alpha\L)}}\frac{\det(\I + \alpha{\featmap}([n])^\transp{\featmap}([n]))}{\det(\I + \alpha\featmap(\coldict)^\transp\W\featmap(\coldict))}\cdot
\end{align*}
To lower bound this quantity we will again upper bound the inverse $\tfrac{\det(\I + \alpha\featmap(\coldict)^\transp\W\featmap(\coldict))}{\det(\I + \alpha{\featmap}([n])^\transp{\featmap}([n]))}$
as follows
\begin{align*}
\tfrac{\det(\I + \alpha\featmap(\coldict)^\transp\W\featmap(\coldict))}{\det(\I + \alpha{\featmap}([n])^\transp{\featmap}([n]))}
&\leq \exp\left\{\tr\left((\alpha\featmap(\coldict)^\transp\W\featmap(\coldict)- \alpha{\featmap}([n])^\transp{\featmap}([n]))(\I + \alpha{\featmap}([n])^\transp{\featmap}([n]))^{-1}\right)\right\}\\
&=\exp\left\{\tr\left(\alpha\featmap(\coldict)^\transp\W\featmap(\coldict)(\I + \alpha{\featmap}([n])^\transp{\featmap}([n]))^{-1}\right) - \deff(\alpha\L)\right\}.
\end{align*}
Inverting the relationship and putting it all together we have
\begin{align*}
&\P(\textit{Acc}_{\sigma} = \text{true})\\
&\geq \exp\left\{\st + r\adeff(\alpha\L) - re^{1/r}\adeff(\alpha\L) + \deff(\alpha\L) - \tr\left(\alpha\featmap(\coldict)^\transp\W\featmap(\coldict)(\I + \alpha{\featmap}([n])^\transp{\featmap}([n]))^{-1}\right) \right\}.
\end{align*}
Using the bound $e^{1/r} \leq 1 + 1/r + 1/r^2$ for $r \geq 1$ we simplify
\begin{align*}
r\adeff(\alpha\L) - re^{1/r}\adeff(\alpha\L)
\geq r\adeff(\alpha\L) - r\adeff(\alpha\L) - \adeff(\alpha\L) - \adeff(\alpha\L)/r
\geq -\adeff(\alpha\L) - \adeff(\alpha\L)/r
\end{align*}
and obtain the final
\begin{align*}
&\P(\textit{Acc}_{\sigma} = \text{true})\\
&\geq \exp\left\{\st - \adeff(\alpha\L) + \deff(\alpha\L) - \tr\left(\alpha\featmap(\coldict)^\transp\W\featmap(\coldict)(\I + \alpha{\featmap}([n])^\transp{\featmap}([n]))^{-1}\right) - \adeff(\alpha\L)/r\right\}.
\end{align*}
Using the definition of $(\varepsilon,\alpha)$-accuracy, we have
\begin{align*}
\adeff(\alpha\L)&= \sum_{i=1}^n\alpha\featmap(i)^\transp(\I + \alpha\featmap(\coldict)^\transp\W\featmap(\coldict))^{-1}\featmap(i)\\
&= \tr\left(\alpha\featmap([n])(\I + \alpha\featmap(\coldict)^\transp\W\featmap(\coldict))^{-1}\featmap([n])^\transp\right)\\
&\leq \tfrac{1}{1 - \varepsilon}\tr\left(\alpha\featmap([n])(\I + \alpha\featmap([n])^\transp\featmap([n]))^{-1}\featmap([n])^\transp\right)\\
&= \tfrac{1}{1 - \varepsilon}\deff(\alpha\L)
= (1 + \tfrac{\varepsilon}{1 - \varepsilon})\deff(\alpha\L),
\end{align*}
and therefore $- \adeff(\alpha\L) + \deff(\alpha\L) \geq \tfrac{\varepsilon}{1 - \varepsilon}\deff(\alpha\L)$. On the other side
\begin{align*}
\tr\left(\alpha\featmap(\coldict)^\transp\W\featmap(\coldict)(\I + \alpha{\featmap}([n])^\transp{\featmap}([n]))^{-1}\right)
&\leq \tfrac{1}{1-\varepsilon}\tr\left(\alpha\featmap(\coldict)^\transp\W\featmap(\coldict)(\I + \alpha\featmap(\coldict)^\transp\W\featmap(\coldict))^{-1}\right)\\
&=\tfrac{1}{1-\varepsilon}\deff(\alpha\Lbh)
= (1 + \tfrac{\varepsilon}{1-\varepsilon})\deff(\alpha\Lbh),
\end{align*}
and therefore $\adeff(\alpha\Lbh) - \tr\left(\alpha\featmap(\coldict)^\transp\W\featmap(\coldict)(\I + \alpha{\featmap}([n])^\transp{\featmap}([n]))^{-1}\right) \geq \tfrac{\varepsilon}{1-\varepsilon}\deff(\alpha\Lbh)$.
Putting it all together, we obtain our result
$\P(\textit{Acc}_{\sigma} = \text{true}) \geq \exp\{-(\varepsilon(\deff(\alpha\L) + \deff(\alpha\Lbh)) + \adeff(\alpha\L)/r))\}$.

To bound $\varepsilon\deff(\alpha\L)$ we simply use the fact that the dictionary
is $1/\deff(\alpha\L)$ accurate. Secondly to bound $\adeff(\alpha\L)/r$ we use
the fact that by \Cref{eq:rls-reformulation} and \Cref{prop:dict-equivalence}
\begin{align*}
\adeff(\alpha\L)
&= \sum_{i=1}^n l_i
= \sum_{i=1}^n\alpha\featmap(i)^\transp(\I + \alpha\featmap(\coldict)^\transp\W\featmap(\coldict))^{-1}\featmap(i)\\
&\leq \frac{1}{1-\varepsilon}\sum_{i=1}^n\alpha\featmap(i)^\transp(\I + \alpha\featmap([n])^\transp\featmap([n]))^{-1}\featmap(i)
= \frac{1}{1-\varepsilon}\sum_{i=1}^n\ell_i(\alpha\L)
= \frac{\deff(\alpha\L)}{1-\varepsilon}.
\end{align*}
Combining this with the fact that $\varepsilon \leq 1/2$
and that $r \geq \deff(\alpha\L)$ we have
that $\adeff(\alpha\L)/r \leq 2$.

Finally, to bound $\varepsilon\deff(\alpha\Lbh)$, first we bound
\begin{align*}
\deff(\alpha\Lbh)
&= \tr(\alpha\W^{1/2}\L_{\coldict}\W^{1/2}(\alpha\W^{1/2}\L_{\coldict}\W^{1/2} + \I)^{-1})\\
&= \tr(\alpha\W^{1/2}\featmap(\coldict)\featmap(\coldict)^\transp\W^{1/2}(\alpha\W^{1/2}\featmap(\coldict)\featmap(\coldict)^\transp\W^{1/2} + \I)^{-1})\\
&= \tr(\alpha\featmap(\coldict)^\transp\W^{1/2}\W^{1/2}\featmap(\coldict)^\transp(\alpha\featmap(\coldict)^\transp\W^{1/2}\W^{1/2}\featmap(\coldict)^\transp + \I)^{-1})\\
&= \tr(\alpha\W\featmap(\coldict)^\transp(\alpha\featmap(\coldict)^\transp\W\featmap(\coldict)^\transp + \I)^{-1}\featmap(\coldict)^\transp)\\
&= \sum_{j=1}^m[\W]_{j,j} \alpha\featmap(\coldict_j)^\transp(\I + \alpha\featmap(\coldict)^\transp\W\featmap(\coldict))^{-1}\featmap(\coldict_j)\\
&\leq \sum_{j=1}^m[\W]_{j,j} \tfrac{1}{1-\varepsilon}\alpha\featmap(\coldict_j)^\transp(\I + \alpha\featmap([n])^\transp\featmap([n]))^{-1}\featmap(\coldict_j)
= \sum_{j=1}^m[\W]_{j,j} \tfrac{1}{1-\varepsilon}\ell_{\coldict_j}(\alpha\L)
\end{align*}
where the last inequality used again \Cref{eq:rls-reformulation} and \Cref{prop:dict-equivalence}.
To continue we have to use the following
result for \bless, the specific dictionary construction algorithm used by
\alphadpp, which follows immediately from
\Cref{prop:bless-r-literature} in \Cref{a:poisson-bless}.
\begin{proposition}[{restate=[name={restated,~first~introduced~on~page~\pageref{prop:bless-specific}}]blessspecific}]\label{prop:bless-specific}
For some $\alpha' \geq \alpha$, let $\coldict$ be a dictionary generated using \pbless
ran with parameter $q \geq 54\kappa^2\tfrac{(2\varepsilon + 1)^{2}}{\varepsilon^2}\log(12n^2/\delta)$
and $\varepsilon \leq \min\{1/2, 1/\deff(\alpha'\L)\}$.
Then w.p.~$1-\delta$
\begin{itemize}
\item the dictionary and weights are $(\varepsilon, \alpha')$-accurate,
\item the weights $\W$ obtained satisfy
$[\W]_{j,j} \leq \max\{\tfrac{1}{1-\varepsilon}\tfrac{1}{q\ell_{\coldict_j}(\alpha'\L)}, 1\}$,
\item the size of the dictionary $m = |\coldict|$ is bounded as
$m/q \leq 2\deff(\alpha'\L)$.
\end{itemize}
\end{proposition}
Applying this to the previous bound, and using the $(1/\deff(\alpha\L), \alpha)$-accuracy, $\varepsilon \leq 1/2$ and the fact that
$\ell_{\coldict_j}(\alpha\L) \leq \ell_{\coldict_j}(\alpha'\L)$ for $\alpha' \geq \alpha$ we obtain
\begin{align*}
\sum_{j=1}^m[\W]_{j,j} \tfrac{1}{1-\varepsilon}\ell_j(\alpha\L)
&\leq
\sum_{j=1}^m
\max\{\tfrac{\deff(\alpha'\L)}{\deff(\alpha'\L) - 1}\tfrac{1}{\deff(\alpha'\L)^2\ell_{\coldict_j}(\alpha'\L)}, 1\}
2\ell_{\coldict_j}(\alpha\L)\\
&= 2\sum_{j=1}^m
\max\{\tfrac{\deff(\alpha'\L)}{\deff(\alpha'\L) - 1}\tfrac{1}{\deff(\alpha'\L)^2}\tfrac{\ell_{\coldict_j}(\alpha\L)}{\ell_{\coldict_j}(\alpha'\L)}, \ell_{\coldict_j}(\alpha\L)\}
\\
&\leq  2\sum_{j=1}^m
\max\{\tfrac{\deff(\alpha'\L)}{\deff(\alpha'\L) - 1}\tfrac{1}{\deff(\alpha'\L)^2}, \ell_{\coldict_j}(\alpha\L)\}\\
&\leq  2\sum_{j=1}^m \left( \tfrac{\deff(\alpha'\L)}{\deff(\alpha'\L) - 1}\tfrac{1}{\deff(\alpha'\L)^2} + \ell_{\coldict_j}(\alpha\L)\right).
\end{align*}
To conclude, we have that since \bless does not include duplicates in $\coldict$,
\begin{align*}
\sum_{j=1}^m \ell_{\coldict_j}(\alpha\L) \leq \sum_{i=1}^n \ell_{i}(\alpha\L)
= \deff(\alpha\L).
\end{align*}
Now using the second result from \Cref{prop:bless-specific} on $m$ we have
\begin{align*}
\sum_{j=1}^m \tfrac{\deff(\alpha'\L)}{\deff(\alpha'\L) - 1}\tfrac{1}{\deff(\alpha'\L)^2}
= m \tfrac{\deff(\alpha'\L)}{\deff(\alpha'\L) - 1}\tfrac{1}{\deff(\alpha'\L)^2}
\leq  2\deff(\alpha'\L) \tfrac{\deff(\alpha'\L)}{\deff(\alpha'\L) - 1}\tfrac{1}{\deff(\alpha'\L)^2}
\leq  2.
\end{align*}
\end{proof}

 \section{Proofs for the binary search algorithm}\label{a:bin-search}
In this section we present omitted proofs for the binary search algorithm.
The key properties of a Poisson Binomial which we will use are
summarized in the following two lemmas.

\begin{lemma}\label{l:pb-classical}
 Let $p: \ZZ_{\geq 0}\rightarrow\R_{\geq 0}$ be a Poisson Binomial distribution. Then:
 \begin{enumerate}
 \item $p$ is unimodal, i.e., if $k^*$ is the mode of $p$, then
$p(1)\leq... \leq p(k^*)\geq p(k^*+1)\geq ...$;
 \item $p$ is log-concave, i.e., $log(p)$ is a concave function over
   the support of $p$;
  \item the median of $p$ is one of  $k^*-1$, $k^*$ and $k^*+1$.
  \end{enumerate}
\end{lemma}
\begin{lemma}[\cite{darroch1964distribution}]\label{prop:pois-bin-dist}
Given a Poisson Binomial with mean
$\bar k$ and mode $k^*$, let $k \defeq \lfloor \bar k \rfloor$. Then:
\begin{equation*}
k^* = \begin{cases}
k & \quad\text{if } \quad k \leq \bar k < k + \frac{1}{k+2}\CommaBin\\
k \;\text{ or }\; k + 1 & \quad\text{if }\quad k + \frac{1}{k+2} \leq
\bar k \leq k + 1 - \frac{1}{n - k + 1}\CommaBin\\
k + 1 & \quad\text{if }\quad k + 1 - \frac{1}{n - k + 1} < \bar k \leq k + 1.
\end{cases}
\end{equation*}
\end{lemma}

Note that these statements are independent of how we break ties in the
definitions of the mode and the median, but for the sake of
concreteness, suppose that we round down when choosing between a pair
of (consecutive) mode/median candidates.

\pbnew*
\begin{proof}[Proof of \Cref{l:pb-new}]
  Let $k^*$ be the mode of $p$ and let $p^*$ denote $p(k^*)$. From
  Lemma \ref{l:pb-classical} it
  follows that $k\neq k^*$. Suppose that $k^*\!>k$ (which implies that
  $k^*\!\geq 2$) and define:
  \begin{align*}
    t \ \defeq\ \min\Big\{ i\in\{1,...,k^*\}\quad \text{subject to}\quad
    p(k^*\!-i)<\frac{p^*}{(1+\beta p^*)^i} \Big\}\CommaBin
  \end{align*}
  where $\beta =2+c/2.5$ is chosen so that the following inequalities
  (used later) hold: (a) \ $\frac1\beta \leq \frac12 -
  \frac c{12}$, (b) \ $\frac 1{(1+\beta)^2}\geq \frac 1{12}$ and (c) \
  $\ee^{\beta}\leq 12$.
If no $i$ exists satisfying the above constraint, then we let
$t=k^*+1$ and use $p(-1)=0$ for convenience. We consider two cases.

\textbf{Case 1:} $t\leq \lceil c/p^*\rceil+1$. Since $p^*\geq \frac c{\sqrt{k^*\!+1}}$,
it follows that $t\leq \big\lceil\sqrt{k^*\!+1}\,\big\rceil+1$. Note that
if $k> k^*-t$ then $k+1\geq
k^*\!+1 - \big\lceil\sqrt{k^*\!+1}\,\big\rceil\geq (k^*\!+1)/3$ and  $p(k)\geq
p^*(1+\beta p^*)^{-1/p^*}>\frac c{\ee^\beta\sqrt{k^*\!+1}}\geq \frac
c{12\sqrt{3(k+1)}}$ which is a contradiction, so we must have $k\leq
k^*-t$. Furthermore, using the definition of $t$ as well as unimodality and
log-concavity of $p$, for any $i\geq t$ we have:
\begin{align*}
  \frac{p(k^*\!-i+1)}{p(k^*\!-i)}\geq
  \frac{p(k^*\!-t+1)}{p(k^*\!-t)}\geq 1 + \beta p^*.
\end{align*}
Thus, $p(k^*\!-i)< \frac{p^*}{(1+\beta p^*)^{i}}$ for all $i\geq t$ and it
follows that:
\begin{align*}
  P_{<k}\leq \sum_{i>t}p(k^*\!-i)\leq
  \frac{p^*}{(1+\beta p^*)^t}\sum_{i\geq 1}\frac1{(1+\beta p^*)^i}\leq
  \frac{p^*}{(1+\beta p^*)^t}\,\frac1{\beta p^*} \leq \frac1\beta\leq
  \frac12-\frac c{12}.
\end{align*}

\textbf{Case 2:} $t>\lceil c/p^*\rceil+1$. This implies that for any
$i\leq \lceil c/p^*\rceil$ we
have $p(k^*\!-i)\geq p^*(1+\beta
p^*)^{-c/p^*}>\frac{c}{12\sqrt{3(k+1)}}> p(k)$ so $k\leq k^*-\lceil c/p^*\rceil-1$. Note that the
median of $p$ is no less than $k^*\!-1$ so:
\begin{align*}
  P_{<k}
  &\leq \underbrace{\sum_{i<k^*\!-1}p(i)}_{\leq 1/2} -
\underbrace{\sum_{i=2}^{\lceil c/p^*\rceil+1}p(k^*\!-i)}_{B}.
\end{align*}
If $p^*\geq \frac c{2\beta}$, then it suffices to note that
\begin{align*}
  B\geq
  p(k^*\!-2)\geq \frac{p^*}{(1+\beta p^*)^2}\geq
  \min\Big\{\frac {c/(2\beta)}{(1+c/2)^2},\, \frac
  1{(1+\beta)^2}\Big\}\geq \frac c{12},
  \end{align*}
whereas if $p^*<\frac c{2\beta}$, then, we have:
\begin{align*}
  B
  &= p^*\bigg(\frac{1-(1+\beta p^*)^{-\lceil c/p^*\rceil-1}}{1-(1+\beta p^*)^{-1}} - 1 -
    (1+\beta p^*)^{-1}\bigg)
= \frac1\beta\Big(1- (1+\beta p^*)^{-\lceil
    c/p^*\rceil}\Big) - \frac{p^*}{1+\beta p^*}
  \\
  &\geq
    \frac1\beta\big(1-2^{-\beta c} - c/2\big)
\geq \frac1\beta\big(1 - (1-\beta c/3) - c/2\big) =\frac c3 - \frac
    c{2\beta} > \frac c{12},
\end{align*}
which completes the proof when $k^*>k$, and the case of $k^*<k$
follows analogously.
\end{proof}

We are now ready to establish the correctness of  the binary search procedure that is used
to prove Lemma \ref{l:binary-search}, with pseudo-code given in
Algorithm \ref{alg:binary-search}. In the proof we will use the
following standard form of the Chernoff bound.
\begin{lemma}[Chernoff bound]\label{l:chernoff}
Let $X_1,...,X_t$ be independent Bernoulli variables and let $\bar X=\frac1t\sum_iX_i$. Then, for any
$0<\epsilon\leq1$, we have:
\begin{align*}
  \Pr\big(|\bar X-\E[\bar X]|\geq \epsilon\cdot\E[\bar X]\big)\leq
  2e^{-\epsilon^2t\,\E[\bar X]/3}.
\end{align*}
\end{lemma}

\binarysearch*
\begin{proof}[Proof of \Cref{l:binary-search}]
Let $\PB(\alpha\L)$ denote the size distribution of $\DPP(\alpha\L)$. Since the
binary search is performed in the log-scale, it takes at most
$O(\log(k\log(\gamma)))$ steps to reduce the interval ratio
$\frac{\alpha_{\max}}{\alpha_{\min}}$ from $\gamma$ to $1+\frac1{(k+3)^2}$.
We first establish
concentration of $\hat P_k$ around its mean
$\E[\hat P_k]=\Pr(|S_1|=k)=p(k)$, where $p$ is the probability
function of $\PB(\bar\alpha\L)$. Define $f(x)=\Pr(\bar
X \geq q/2)$ where $\bar X=\frac1t \sum_{i=1}^t X_i$
and $X_i$ are drawn i.i.d.~from
$\mathrm{Bernoulli}(x)$, with $q=\frac{c}{12\sqrt{3(k+1)}}$. Lemma
\ref{l:chernoff} implies that, choosing a sufficiently large constant
$C$ in \Cref{alg:binary-search}, we have:
\begin{align*}
\max\big\{f(q/4),\,1-f(q)\big\} \leq 2\ee^{-t q/12}\leq \frac{\delta}{4s^2}\CommaBin
\end{align*}
where $s$ is the number of the current branching step.
Note that if $p(k)>q$ then $\Pr(\hat P_k<q/2) \leq 1-f(q)\leq
\delta/(4s^2)$ whereas if $p(k)<q/4$, then $\Pr(\hat P_k\geq q/2)\leq
f(q/4)\leq\delta/(4s^2)$, so putting this together we conclude that:
\begin{align*}
\Pr\Big(&\big(\hat P_k\geq \tfrac q2 \Rightarrow p(k)\geq \tfrac q4\big)\wedge
  \big(\hat P_k<\tfrac q2 \Rightarrow p(k)<q\big)\Big)
  \\
&=1 - \Pr\Big(\big(\hat P_k\geq \tfrac q2\wedge p(k)<\tfrac q4\big)
\vee \big(\hat P_k<\tfrac q2\wedge p(k)\geq q\big)\Big)\geq 1-\frac{\delta}{4s^2}\cdot
\end{align*}
Thus, conditioning on the above high probability event ensures that when the
\textbf{if} statement in Line \ref{line:if2} of Algorithm \ref{alg:binary-search} succeeds
then $\hat\alpha$ satisfies the condition from Lemma~\ref{l:binary-search} because $p(k)\geq
q/4=\Omega(\frac1{\sqrt k})$, and when
the if statement fails, then the assumption of Lemma~\ref{l:pb-new} is
satisfied because $p(k)<q$.

We now move on to the branching step of the binary search (Line~\ref{line:branch}). Our
assumptions ensure that the initial interval
$(\alpha_{\min},\alpha_{\max})$ contains an $\alpha_\star$ such that $k$
is the mode of $\PB(\alpha_\star\L)$. Our goal is to show that the
branching step preserves this invariant throughout the procedure. As
discussed above, when entering the branching step, with high
probability we have $p(k)<q$, so that we can use Lemma \ref{l:pb-new}.
Note that $\E[\hat P_{<k}]=P_{<k}$ and
$\E[\hat P_{>k}]=P_{>k}$, as defined in
the lemma, and the goal of the branching statement is to
determine whether $P_{<k}>P_{<k}$, since that tells us on which side of $k$
is the mode of $\PB(\bar\alpha K)$. Conditioned
on a high probability event, we know that either
$P_{<k}\leq \frac12 - \tfrac c{12}$ or
$P_{>k}\leq\frac12-\tfrac{c}{12}$. Suppose the former holds. Then, we have:
\begin{align*}
P_{>k} = 1 -(P_{<k}+p(k))\geq 1 - (\tfrac12 - \tfrac c{12} +
  \tfrac{c}{12\sqrt{3}})\geq \tfrac12 + \tfrac c{30},
\end{align*}
and an analogous bound follows for $P_{<k}$ in the latter case.
If $P_{>k}\geq \frac12+\frac c{30}$ (call it event $E$), then we can once again apply
Lemma~\ref{l:chernoff} to show that (for a sufficiently large constant $C$),
\begin{align*}
  \Pr\big(\hat P_{<k}>\hat P_{>k}\mid E\big)
  & \geq\Pr\big(\hat P_{<k}\geq \tfrac12\mid E\big)
\geq  \Pr\big(|\hat P_{<k}-\E[P_{<k}]| <\tfrac c{30}\mid E\big)
  \\
  &\geq1-
  2\exp\big\{-(\tfrac{c}{30})^2\tfrac12 t/3\big\}\geq1- \delta/(4s^2),
\end{align*}
and an analogous claim follows when
$P_{>k}\leq\frac12-\frac{c}{12}$. Conditioning on this high
probability event implies
(via Lemma~\ref{l:pb-new}) that the interval constructed after
branching still satisfies the invariant. A union bound implies that
the probability that any of the events we have conditioned on fails
(throughout the algorithm) is bounded by
$\sum_{s\geq 1}\frac{2\delta}{4 s^2}\leq\delta$. Thus, with probability
$1-\delta$ the last interval used in the search will still satisfy
the invariant. It remains to show that when the \textbf{if} statement in Line~\ref{line:if1}
succeeds then either $\alpha_{\min}$ or $\alpha_{\max}$ satisfies
the claim of Lemma \ref{l:binary-search}.
To that end, since $k\geq\lfloor\deff(\alpha_{\min}\L)\rfloor$, we have:
\begin{align*}
  \deff(\alpha_{\max}\L)
  &< \deff\big((1+\tfrac1{(k+3)^2})\alpha_{\min} \L\big)
\leq \big(1+\tfrac1{(k+3) ^2}\big)\deff(\alpha_{\min}\L)\
  \\
  &\leq
  \deff(\alpha_{\min}\L)+ \frac{1}{\lfloor\deff(\alpha_{\min}\L)\rfloor+3}\cdot
\end{align*}
Now, there are two cases. Either
$\lfloor\deff(\alpha_{\max}\L)\rfloor=\lfloor\deff(\alpha_{\min}\L)\rfloor$,
in which case Lemma \ref{prop:pois-bin-dist} immediately implies that there are at
most two possible modes of the Poisson
Binomial $\PB(\alpha \L)$ among all values of
$\alpha\in[\alpha_{\min},\alpha_{\max}]$, and they
must be achieved by $\alpha_{\min}$ and by $\alpha_{\max}$. If
$\lfloor\deff(\alpha_{\max}\L)\rfloor=\lfloor\deff(\alpha_{\min}\L)\rfloor+1$,
then the same conclusion is reached by observing that:
\begin{align*}
  \deff(\alpha_{\max}\L) \leq  \lfloor\deff(\alpha_{\max}\L)\rfloor +
  \frac1{\lfloor\deff(\alpha_{\min}\L)\rfloor+3}\leq
  \lfloor\deff(\alpha_{\max}\L)\rfloor
  +\frac1{\lfloor\deff(\alpha_{\max}\L)\rfloor+2},
\end{align*}
so, by Lemma \ref{prop:pois-bin-dist}, the mode of
$\PB(\alpha_{\max}\L)$ must be $\lfloor\deff(\alpha_{\max}\L)\rfloor$, and once
again there are only two possible modes in the interval $\alpha\in[\alpha_{\min},\alpha_{\max}]$.
With high probability, one of
these modes must be $k$, which completes the proof.
\end{proof}

 \section{\pbless algorithm}\label{a:poisson-bless}
In this section we present the omitted \pbless algorithm with proofs of its
accuracy and efficiency. For simplicity, in the entirety
of this section we will assume that $k \geq 2$. Note that this can be relaxed,
at the only cost of slightly more complex constants (\eg $\alpha_{\tinit} = \max\{k-1,1\}/\Ltrace)$ instead of $\alpha_{\tinit} = (k-1)/\Ltrace)$.
Moreover, the case $k =1$ is qualitatively different, as in a $1$-DPP
the marginal and joint distribution coincide, making it much simpler to sample from.

\subsection{\bless}
\begin{algorithm}[t]
  \caption{\bless (rejection-based version)}
\label{alg:bless-r}
\begin{algorithmic}[1]
\renewcommand{\algorithmicrequire}{\textbf{Input:}}
\REQUIRE $\L\in\R^{n\times n}$,
\ $q>0$, $k$, $\alpha_{\max}$
\STATE Initialize
$i=0$, $\alpha^0 = 1/\Ltrace$,
$\edeff(\alpha^0\Lb) = \tfrac{1}{2}(k-1)$
\STATE Initialize $\coldict^0$ by sampling $q\alpha^0n\kappa^2$ elements $\coldict^0 \simiid (1/n, \ldots, 1/n)$ and weight $w_j^0 = 1/(q\alpha^0\kappa^2)$.
\FOR{$i = \{1, \ldots, \lceil\log_2(\alpha_{\max}/\alpha^0)\rceil\}$}
\STATE Set $\alpha^i = 2\alpha^{i-1}$, $b^i = \min\{q\alpha^{i}\kappa^2,1\}$
\FOR{$j = \{1, \dots, n\}$}
\STATE Sample $u^i_j \sim \Bernoulli(b^i)$
\IF{$u^i_j = 1$}
\STATE Compute $l_{_j}^i$ using \Cref{eq:rls-approx} and $\coldict^{i-1}$
\STATE Sample $z_j^i \sim \Bernoulli(\min\{ql_{j}, b^i\}/b^i)$
\ENDIF
\ENDFOR
\STATE Set $\sigma^i = \{j \in [n]: z_j^i = 1\}$, $\coldict^i = \sigma^i$, $w_j^i = 1/\min\{ql_{\sigma^i_j}, b^i\}$
\ENDFOR
\RETURN $\coldict^{\alpha^{\lceil\log_2(\alpha_{\max}/\alpha^0)\rceil}}$
\end{algorithmic}
\end{algorithm}

We begin by reporting the \bless algorithm \cite{NIPS2018_7810}
and several of its properties.
Note that \bless was originally introduced as a ridge leverage score
(RLS) sampling algorithm. However in the context of DPPs the RLS of an item
coincides exactly with its marginal inclusion probability, \ie $\ell_i(\L)$
is the RLS of the $i$-th item. Therefore we can leverage any RLS sampler
both to generate dictionaries as well as RLS estimate for \alphadpp. We choose
to use \bless as a starting point because, to our knowledge, it is the only rescaling-aware
RLS sampler existing in the literature. We report \bless, in its rejection sampling version,
in full in \Cref{alg:bless-r} with the only notational difference of using
a rescaling $\alpha \leq 1$ rather than a regularization $\lambda$, with a conversion
$\alpha \approx 1/(\lambda n)$ between the two.

\begin{proposition}[Thm.\,1 by \citet{NIPS2018_7810}]\label{prop:bless-r-literature}
For some $\alpha' \geq \alpha$, let $\coldict$ be a dictionary generated using \bless
ran with parameter $q \geq 54\kappa^2\tfrac{(2\varepsilon + 1)^{2}}{\varepsilon^2}\log(12n^2/\delta)$.
Then w.p.~$1-\delta$ for all $i,$
  {
  \setlength{\itemsep}{-2pt}
\begin{itemize}[leftmargin=*]
\item the dictionary $\coldict^i$ and weights are $(\varepsilon, \alpha^i)$-accurate,
\item the approximate marginals $l_j$ computed using $\coldict^i$ satisfy
$\frac{1}{1+\varepsilon} \ell_j(\alpha^i) \leq l_j \leq \frac{1}{1-\varepsilon} \ell_j(\alpha^i)$.
\item the size of the dictionary $m^i = |\coldict^i|$ is bounded as
$\deff(\alpha^i\L)/2 \leq m/q \leq 2\deff(\alpha^i\L)$,
\end{itemize}
}
and the algorithm runs in $\cO\left((\min\{\alpha_{\max} n \kappa^2, 1\}\deff(\alpha_{\max}\L)^2\log(n/\delta)^3)\log(\alpha_{\max}\Ltrace)\right)$ time.
\end{proposition}
Note that all results presented in \Cref{prop:bless-r-literature} are only
reformulations of Theorem 1 from \cite{NIPS2018_7810}. The only exception
is the lower bound $m/q \geq \deff(\alpha^i\L)/2$, since the original
\bless analysis was only interested in showing that $m/q \leq \deff(\alpha^i\L)/2$.
However, the same concentration argument of Lemma 6 in \cite{NIPS2018_7810}
also holds for the lower bound we report here.

\subsection{Modification to \bless}
In order to use \bless in our approach for DPP sampling, we need to make a few
modifications. Compared to \bless, our \pbless (\Cref{alg:adjusted-bless}):
\begin{itemize}[leftmargin=*]
\item automatically computes an appropriate $\alpha_{\max}$ rather than taking it as input;
\item introduces a novel $\alpha_{\tinit}$ to initialize $\alpha^0$ that both takes into account the desired DPP size $k$ and is a valid lower bound for the interval search;
\item automatically computes an appropriate $\alpha_{\min}$ rather than setting $\alpha_{\min} = \alpha^{0}$;
\item uses the last $\deff(\alpha_{\max}\L)$ estimate to generate a dictionary
$\coldict^{\alpha_{\max}}$ that is guaranteed to be $(1/\deff(\alpha_{\max}\L), \alpha_{\max})$-accurate.
\end{itemize}
\begin{algorithm}[t]
  \caption{\bless modified to compute the search interval (\pbless)}
\label{alg:adjusted-bless}
\begin{algorithmic}[1]
\renewcommand{\algorithmicrequire}{\textbf{Input:}}
\REQUIRE $\L\in\R^{n\times n}$,
\ $q>0$, $k$
\STATE Initialize
$i=0$, $\alpha^0 = \alpha_{\tinit} =  (k-1)/(n\kappa^2)$,
$\edeff(\alpha^0\Lb) = \tfrac{1}{2}(k-1)$
\STATE Initialize $\coldict^0$ by sampling $q\alpha^0n\kappa^2$ elements $\coldict^0 \simiid (1/n, \ldots, 1/n)$ and weight $w_j^0 = 1/(q\alpha^0\kappa^2)$.
\WHILE{$\edeff(\alpha^i\Lb) \leq 2(k + 2)$\label{line:pbless-exit}}
\STATE Set $i = i+1$, $\alpha^i = 2\alpha^{i-1}$, $\alpha_{\max} = \alpha^{i}$, $b^i = \min\{q\alpha^{i}\kappa^2,1\}$
\FOR{$j = \{1, \dots, n\}$}
\STATE Sample $u^i_j \sim \Bernoulli(b^i)$
\IF{$u^i_j = 1$}
\STATE Compute $l_{_j}^i$ using \Cref{eq:rls-approx} and $\coldict^{i-1}$
\STATE Sample $z_j^i \sim \Bernoulli(\min\{ql_{j}, b^i\}/b^i)$
\rlap{
\hspace*{3.35cm}\smash{$\left.\begin{array}{@{}c@{}}\\{}\\{}\\{}\\{}\\{}\\{}\\{}\\{}\\{}\\{}\\{}\\{}\\{}\\{}\end{array}\color{black}\right\}
 \color{black}\begin{tabular}{l}Computing\\approximate RLS.\end{tabular}$}
\vspace*{-.2cm}
}
\ENDIF
\ENDFOR
\STATE Set $\sigma^i = \{j \in [n]: z_j^i = 1\}$, $\coldict^i = \sigma^i$, $w_j^i = 1/\min\{ql_{\sigma^i_j}, 1\}$
\STATE set $\edeff(\alpha^i\Lb) = |\coldict^i|/q$
\IF{$\edeff(\alpha^{i-1}\Lb) \leq \tfrac{1}{2}(k - 1)$ and $\edeff(\alpha^{i}\Lb) > \tfrac{1}{2}(k - 1)$\label{line:pbless-min}}
\STATE Set $\alpha_{\min} = \alpha^{i-1}$
\ENDIF
\ENDWHILE
\STATE Set $\coldict^{\alpha_{\max}} = \emptyset$, $q' = q\edeff(\alpha^i\Lb)^2$, $b^{\max} = \min\{q'\alpha^i\kappa^2,1\}$
\FOR{$j = \{1, \dots, n\}$}
\STATE Sample $u^{\max}_j \sim \Bernoulli(b^{\max})$
\IF{$u^{\max}_j = 1$}
\STATE Compute $l_{_j}^{\max}$ using \Cref{eq:rls-approx} and $\coldict^{i}$\hspace*{3.7cm}
\rlap{\smash{$\left.\begin{array}{@{}c@{}}\\{}\\{}\\{}\\{}\\{}\\{}\\{}\\{}\\{}\end{array}\color{black}\right\}
 \color{black}\begin{tabular}{l}Final dictionary\\construction.\end{tabular}$}}
\STATE Sample $z_j^{\max} \sim \Bernoulli(\min\{q'l_{j}^{\max}, b^{\max}\}/b^{\max})$
\STATE If $z_{j}^{\max} = 1$, add $j$ to $\coldict^{\alpha_{\max}}$ with weight $w_j^{\alpha_{\max}} = \tfrac{1}{\min\{q'l_{j}^{\max}, b^{\max}\}}$
\ENDIF
\ENDFOR
\RETURN $\alpha_{\min}$, $\alpha_{\max}$, $\coldict^{\alpha_{\max}}$
\end{algorithmic}
\end{algorithm}

\lambdalowerbound*
\begin{proof}[Proof of \Cref{lem:lambda-lower-bound}]
From the definition
$\deff(\alpha\Lb)
= \tr(\alpha\Lb(\alpha\Lb + \I)^{-1})$.
Then the first half comes from
\begin{align*}
\tr(\alpha\Lb(\alpha\Lb + \I)^{-1}) \leq \tr(\alpha\Lb(\I)^{-1}) = \alpha\tr(\Lb),
\end{align*}
while for the second half we have
\begin{align*}
\tr(\alpha\Lb(\alpha\Lb + \I)^{-1}) \geq \alpha\tr(\Lb(\Lb + \I)^{-1})
= \alpha\deff(\Lb).
\end{align*}
\end{proof}

\algovalidinterval*
\begin{proof}[Proof of \Cref{lem:algo-valid-interval}] Throughout the proof we will
make use of \Cref{prop:bless-r-literature}, in particular that
$\tfrac{1}{2} \deff(\alpha^i\Lb) \leq \edeff(\alpha^i\Lb)\leq {2} \deff(\alpha^i\Lb)$.
Note that by inverting the relationship we also have
the reciprocal guarantee
$\tfrac{1}{2} \edeff(\alpha^i\Lb) \leq \deff(\alpha^i\Lb)\leq {2} \edeff(\alpha^i\Lb)$.

\textbf{Claim (1): size of the interval.}
Applying \Cref{lem:lambda-lower-bound} we have that
$\alpha_{\max} \leq \deff(\alpha_{\max}\Lb)/\deff(\Lb)$. We need now to
further upper bound $\deff(\alpha_{\max}\Lb)$ \pbless's terminating condition (\Cref{line:pbless-exit})
only guarantees the lower bound $\edeff(\alpha_{\max}\Lb) \geq 2(k+2)$.
To this end we will use a property of RLS (see Lemma~3 from \cite{NIPS2018_7810})
that says that if $\alpha^{i} > \alpha^{i-1}$
then $\deff(\alpha^i\Lb) \leq \tfrac{\alpha^i}{\alpha^{i-1}}\deff(\alpha^{i-1}\Lb)$.
In our case, $\alpha^{i}/\alpha^{i-1} = 2$ and $\deff(\alpha^i\Lb) \leq 2\deff(\alpha^{i-1}\Lb)$.
Now, let $i$ be the index before the
loop exit condition in \Cref{alg:adjusted-bless} is satisfied (\ie $\alpha_{\max} = \alpha^{i+1}$).
Then we have $\edeff(\alpha^{i}\Lb) \leq 2(k+2)$, using \Cref{prop:bless-r-literature} we further bound $\deff(\alpha^i\Lb) \leq 4(k+2)$, which implies
that $\deff(\alpha^{i+1}\Lb) = \deff(\alpha_{\max}\Lb) \leq 8(k+2)$.
Going back to our bound we obtain $\alpha_{\max} \leq \deff(\alpha_{\max}\Lb)/\deff(\Lb) \leq 8(k+2)/\deff(\Lb)$.

The side of $\alpha_{\min}$ is much simpler. From \Cref{lem:lambda-lower-bound} we have that $\alpha_{\min} \geq \deff(\alpha_{\min}\Lb)/\Ltrace$,
and from the algorithm we know that $\edeff(\alpha_{\min}\Lb) \geq \tfrac{1}{2}(k-1)$.
Combining this with \Cref{prop:bless-r-literature} we get
\begin{align*}
\alpha_{\min} \geq \deff(\alpha_{\min}\Lb)/\Ltrace
\geq \tfrac{1}{2}\edeff(\alpha_{\min}\Lb)/\Ltrace
\geq \tfrac{1}{4}(k-1)/\Ltrace.
\end{align*}

\textbf{Claim (2): validity of the interval.}
To begin, remember from \Cref{prop:pois-bin-dist} that
the mode $m_{\alpha}$ of the sample size of $\DPP(\alpha\L)$ is bounded by $\lfloor\deff(\alpha\L)\rfloor \leq m_{\alpha} \leq \lfloor\deff(\alpha\L)\rfloor + 1$.
To guarantee the validity of our interval, we show that $m_{\alpha_{\min}} \leq k$, and $m_{\alpha_{\max}} \geq k + 1$.
Due to the monotonicity of the mode of a Poisson Binomial distribution (see \Cref{l:pb-classical})
we have therefore that starting from $m_{\alpha_{\min}}$ the mode increases with $\alpha$,
until it reaches $k$ for some $\alpha_{\star} \in [\alpha_{\min}, \alpha_{\max}]$,
and then continue increasing until it reaches $k + 1 \leq m_{\alpha_{\max}}$.

Concretely, we have that
\begin{align*}
m_{\alpha_{\min}}
\stackrel{\text{\Cref{prop:pois-bin-dist}}}{\leq} \deff(\alpha_{\min}\Lb) + 1
\stackrel{\text{\Cref{prop:bless-r-literature}}}{\leq} 2\cdot\edeff(\alpha_{\min}\Lb) + 1
< 2\cdot\tfrac{1}{2}(k - 1) + 1
= k,
\end{align*}
where the last inequality is due to the condition from \Cref{line:pbless-min} in \pbless. Similarly
\begin{align*}
m_{\alpha_{\max}}
\stackrel{\text{\Cref{prop:pois-bin-dist}}}{\geq} \deff(\alpha_{\max}\Lb) - 1
\stackrel{\text{\Cref{prop:bless-r-literature}}}{\geq} \tfrac{1}{2}\cdot\edeff(\alpha_{\max}\Lb) - 1
> \tfrac{1}{2}\cdot 2(k + 2) - 1
= k + 1,
\end{align*}
where this time the last inequality is due to the condition from \Cref{line:pbless-exit} in \pbless.
Finally, we have to guarantee that $\alpha_{\tinit}$ is also a valid lower bound,
or we will never able to correctly set $\alpha_{\min}$. This is easy to show
using \Cref{lem:lambda-lower-bound}
\begin{align*}
m_{\alpha_{\tinit}} \leq \deff(\alpha_{\tinit}\L) + 1
\stackrel{\text{\Cref{lem:lambda-lower-bound}}}{\leq} \alpha_{\tinit}\Ltrace + 1
= \tfrac{\Ltrace}{n\kappa^2}(k-1) + 1
\leq k-1 + 1
= k,
\end{align*}
making it a valid initialization for the lower bound.

\textbf{Claim (3): quality of $\coldict^{\max}$.} At the end of the main loop,
due to \Cref{prop:bless-r-literature} we have that $\edeff(\alpha^i\L) \geq \tfrac{1}{2}\deff(\alpha^i\L)$, and that
since $\alpha^i = 2\alpha^{i-1} = 2\alpha_{\max}$, $\deff(\alpha^i\L) \geq \deff(\alpha_{\max}\L)$.
Therefore, setting $q' = 4\edeff(\alpha^i\L)^2q$ is sufficient to invoke \Cref{prop:bless-r-literature}
with $\varepsilon = 1/\deff(\alpha_{\max}\L)$ and obtain an $(1/\deff(\alpha_{\max}), \alpha_{\max})$-accurate dictionary.
Moreover, it is easy to see that for any $\alpha' \geq \alpha$ and $\varepsilon' \leq \varepsilon$, an $(\varepsilon',\alpha')$-accurate dictionary
is also an $(\varepsilon, \alpha)$-accurate dictionary (see \Cref{prop:dict-equivalence}).
Since $\alpha_{\max} \geq \alpha$ for the whole duration of the binary search,
and therefore $\deff(\alpha_{\max}\L) \geq \deff(\alpha\L)$, our $\coldict^{\max}$ dictionary is sufficiently
accurate for the whole duration of the binary search.
\end{proof}

 \section{Additional experimental details}\label{a:experiments}
\newcommand{\qbless}{q_{\bless}}
\newcommand{\qdpp}{q_{\text{dpp}}}

\begin{figure*}
\noindent\begin{minipage}{0.49\textwidth}
\centering
\includegraphics[width=\textwidth]{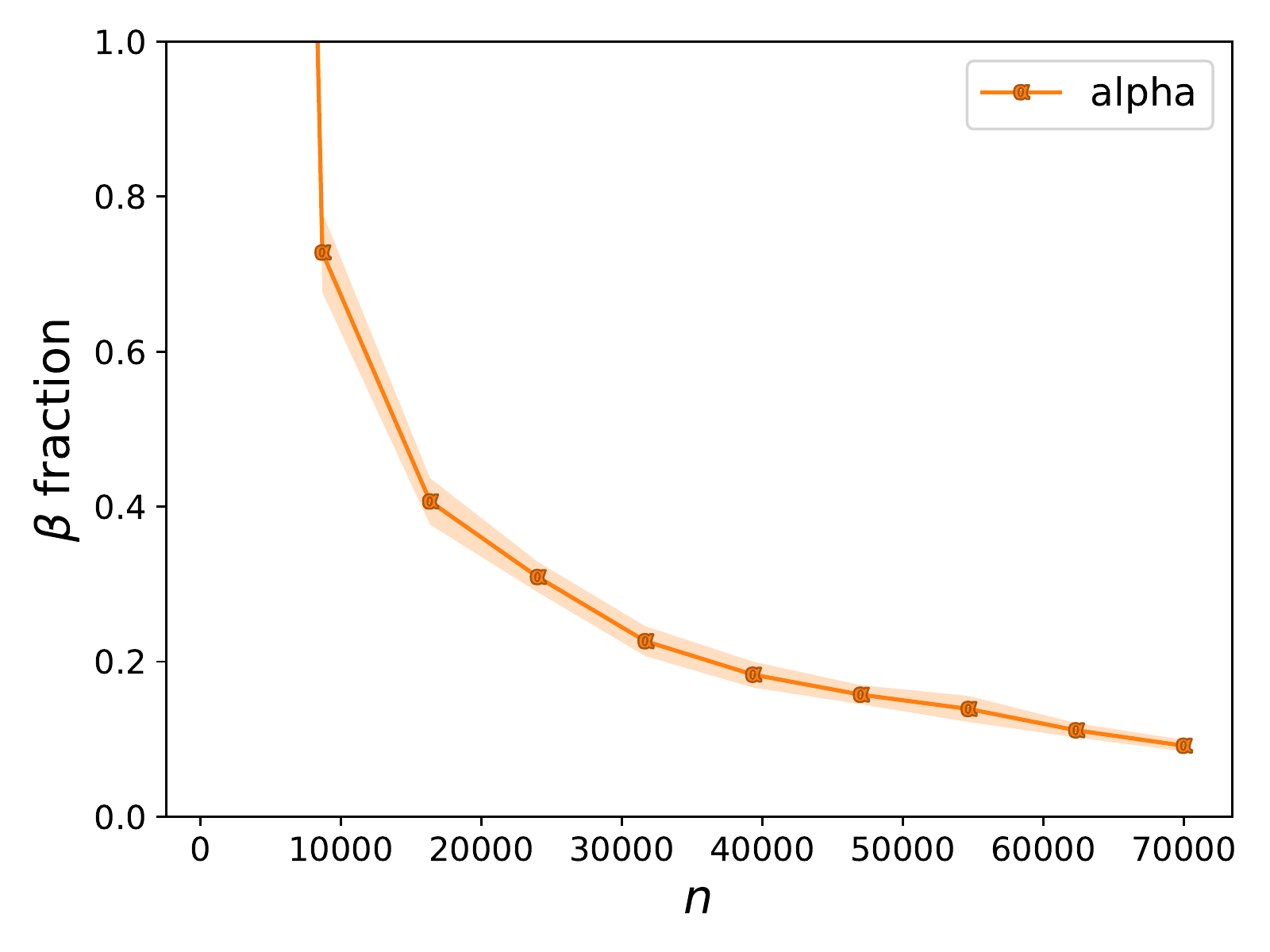}
\vspace{-2mm}
\captionof{figure}{Fraction of items observed by
  \alphadpp on the small scale experiment.}\label{fig:fraction-small-scale}
\end{minipage}\hfill
\begin{minipage}{0.49\textwidth}
\centering
\includegraphics[width=\textwidth]{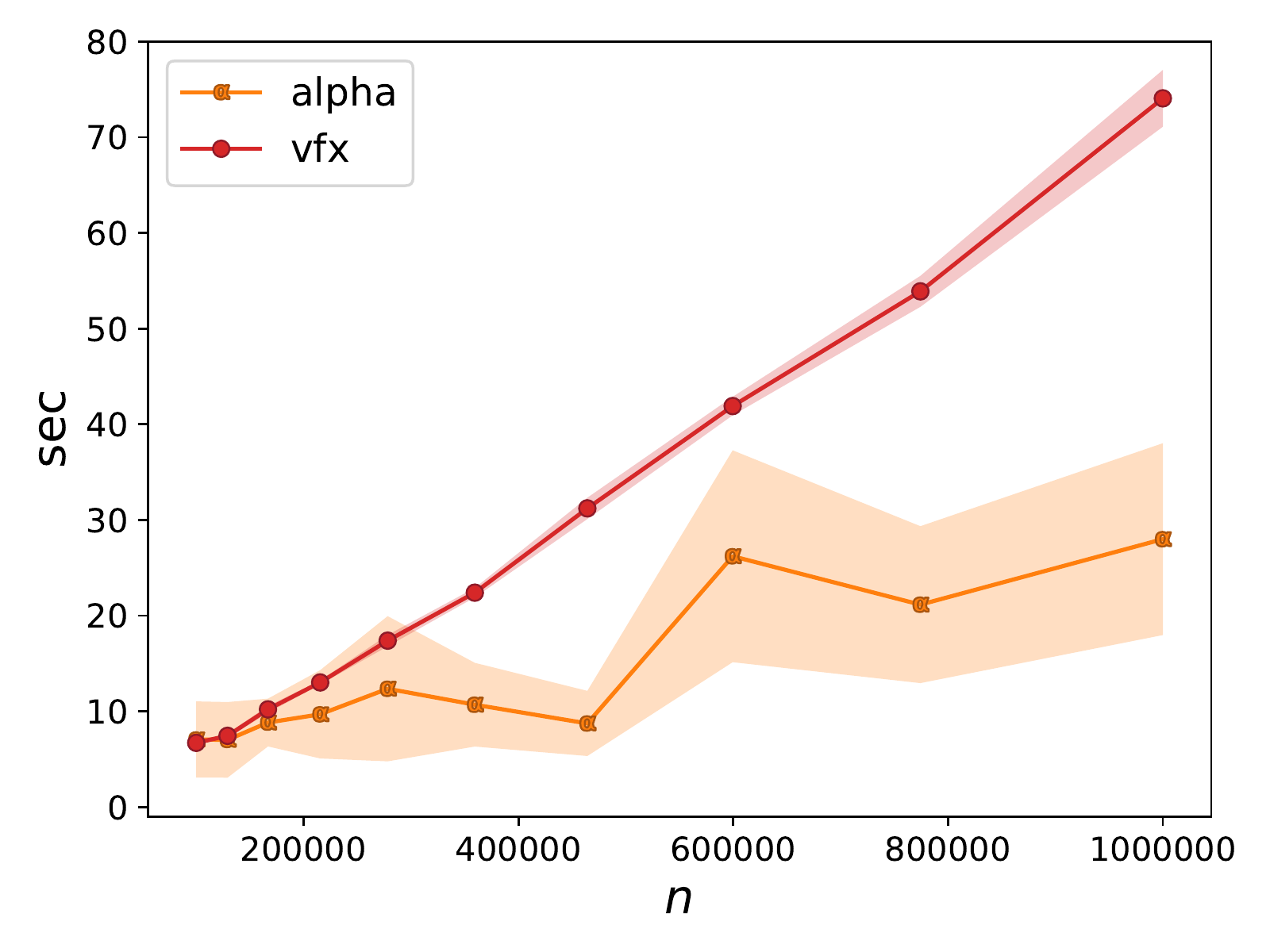}
\vspace{-2mm}
\captionof{figure}{Large scale experiment using linear similarity.}\label{fig:runtime-linear}
\end{minipage}\hfill
\end{figure*}

Both \dppvfx and \alphadpp rely on \bless or \pbless to generate their input
dictionaries. For this preprocessing phase, the major hyperparameters to tune are
$\qbless$ and $\qdpp$, \ie the $q$ and $q'$ parameters indicated in
\Cref{alg:adjusted-bless}.
\footnote{Following \texttt{DPPy}'s API, these hyperparameters are denoted as \texttt{rls\_oversample\_bless} and \texttt{rls\_oversample\_dppvfx} in our code.}

Note that theory suggests to set $\qbless \approx \cO(\log(n))$ and $\qdpp \approx \cO(\deff(\alpha\L)^2)$,
but they can be freely tuned
since both \alphadpp and \dppvfx remain exact samplers for any hyperparameter
choice. However, $\qbless$ and $\qdpp$ do impact acceptance rate and runtime,
and even more importantly too low values can result in empty dictionaries
which force the algorithm to be stopped.

In our case, we start with $\qbless = 2$ and $\qdpp = 2$, and increase them until
the \texttt{DPPy} implementation does not return an empty dictionary.
We also keep the same value for \alphadpp and \dppvfx so that for similar $\alpha$
they operate with similarly accurate and large dictionaries.
The final values are $\qbless = 5$ and $\qdpp = 10$ for the small scale experiment
(\Cref{fig:small-scale}), and $\qbless = 4$ and $\qdpp = 5$ for the large scale experiment (\Cref{fig:large-scale}).

For completeness, in addition to the fraction of observed items in the large scale experiment (\Cref{fig:rejections}),
we also report the fraction of observed items in the small scale experiment (\Cref{fig:fraction-small-scale}).
We note that, for the small scale experiment, until $n$ exceeds 10000,
\alphadpp is still observing all items,
and only when the item collection becomes sufficiently large uniform
sampling starts to play a role.

Finally, we report another experiment taken directly from the benchmark of
\citet{derezinski2019exact} where a linear similarity is used instead of
\texttt{rbf} similarity. We see that in this setting $\deff(\L)$ grows
slower with $n$, since the similarity/kernel is less expressive. As a consequence
the gap between \alphadpp and \dppvfx (\ie the advantage of using uniform intermediate sampling)
is reduced, but remains impactful.
 \section{Miscellaneous proofs}\label{a:misc-proofs}
In this section we present omitted miscellaneous facts and proofs for completeness.
\begin{definition}\label{d:rls}
Given a psd matrix $\L$, its $i$th ridge leverage score (RLS)
$\ell_i(\L)$ is the $i$th diagonal entry of
$\L(\I+\L)^{-1}$. The
sum $\sum_{i=1}^n\ell_i(\L) = \deff(\L)$ of the RLSs is equal to the effective dimension
of $\L$.
\end{definition}
\begin{definition}[\cite{ridge-leverage-scores,calandriello_disqueak_2017}]
A dictionary $\coldict$ and its associated weighting matrix $\W$ are $(\varepsilon,\alpha)$-accurate
if
$\normsmall{\alpha\L(\I + \alpha\L)^{-1}(\I - \wb{\W})} \leq \varepsilon$,
where $\wb{\W} \in \R^{n \times n}$ is diagonal with $\wb{\W}_{i,i} = \sum_{j=1}^m w_j \indfunc\{\coldict_j = i\}$.
\end{definition}

\begin{proposition}[\cite{ridge-leverage-scores,calandriello_disqueak_2017}]\label{prop:dict-equivalence}
A dictionary $\coldict$ and its associated weighting matrix $\W$ are
$(\varepsilon,\alpha)$-accurate
if
\begin{align*}
\normsmall{(\I + \alpha\featmap([n])^\transp\featmap([n]))^{-1/2}(\alpha\featmap([n])^\transp\featmap([n]) - \alpha\featmap(\coldict)^\transp\W\featmap(\coldict))(\I + \alpha\featmap([n])^\transp\featmap([n]))^{-1/2}} \leq \varepsilon,
\end{align*}
or equivalently
\begin{align*}
\normsmall{(\I/\alpha + \featmap([n])^\transp\featmap([n]))^{-1/2}(\featmap([n])^\transp\featmap([n]) - \featmap(\coldict)^\transp\W\featmap(\coldict))(\I/\alpha + \featmap([n])^\transp\featmap([n]))^{-1/2}} \leq \varepsilon,
\end{align*}
or yet equivalently
\begin{align*}
(1-\varepsilon)(\I/\alpha + \featmap([n])^\transp\featmap([n])) \preceq \I/\alpha + \featmap(\coldict)^\transp\W\featmap(\coldict)
 \preceq (1+\varepsilon)(\I/\alpha + \featmap([n])^\transp\featmap([n])).
\end{align*}
\end{proposition}
Note that using \Cref{prop:dict-equivalence} it is easy to see that for any $\alpha' \geq \alpha$ and $\varepsilon' \leq \varepsilon$, an $(\varepsilon',\alpha')$-accurate dictionary is also an $(\varepsilon,\alpha)$-accurate dictionary since $\I/\alpha' \preceq \I/\alpha$ and therefore
\begin{align*}
&\normsmall{(\I/\alpha + \featmap([n])^\transp\featmap([n]))^{-1/2}(\featmap([n])^\transp\featmap([n]) - \featmap(\coldict)^\transp\W\featmap(\coldict))(\I/\alpha + \featmap([n])^\transp\featmap([n]))^{-1/2}}\\
&\leq
\normsmall{(\I/\alpha' + \featmap([n])^\transp\featmap([n]))^{-1/2}(\featmap([n])^\transp\featmap([n]) - \featmap(\coldict)^\transp\W\featmap(\coldict))(\I/\alpha' + \featmap([n])^\transp\featmap([n]))^{-1/2}}\\
&\leq \varepsilon'
\leq \varepsilon.
\end{align*}
Moreover, using basic algebraic manipulation we can see that for any matrix/operator $\A$ we have
\begin{align*}
(\I + \A\A^\transp)^{-1} = \I - \A(\I + \A^\transp\A)^{-1}\A^\transp,
\end{align*}
which applied to $\A = \sqrt{\alpha}\W^{1/2}\featmap(\coldict)$ gives us the following reformulation from \cite{calandriello_disqueak_2017, NIPS2018_7810}:
\begin{align*}
l_i &=  \alpha[\L - \L_{[n],\coldict}^\transp(\alpha\L_{\coldict,\coldict} + \W^{-1})^{-1}\L_{[n],\coldict}]_{i,i}\\
&=  \alpha[\featmap([n])\featmap([n])^\transp - \alpha\featmap([n])\featmap(\coldict)^\transp(\alpha\featmap(\coldict)\featmap(\coldict)^\transp + \W^{-1})^{-1}\featmap(\coldict)\featmap([n])^\transp]_{i,i}\\
&=  \alpha[\featmap([n])\left(\I  - \alpha\featmap(\coldict)^\transp(\alpha\featmap(\coldict)\featmap(\coldict)^\transp + \W^{-1})^{-1}\featmap(\coldict)\right)\featmap([n])^\transp]_{i,i}\\
&=  \alpha[\featmap([n])\left(\I  - \alpha\featmap(\coldict)^\transp\W^{1/2}(\alpha\featmap(\coldict)\W\featmap(\coldict)^\transp + \I)^{-1}\W^{1/2}\featmap(\coldict)\right)\featmap([n])^\transp]_{i,i}\\
&= \alpha[\featmap([n])(\I + \alpha\featmap(\coldict)^\transp\W\featmap(\coldict))^{-1}\featmap([n])^\transp]_{i,i}\\
&= \alpha\featmap(i)^\transp(\I + \alpha\featmap(\coldict)^\transp\W\featmap(\coldict))^{-1}\featmap(i).
\end{align*}
Applying \Cref{prop:dict-equivalence} to the reformulation it is easy to see that
\begin{align*}
&\alpha\featmap(i)^\transp(\I + \alpha\featmap(\coldict)^\transp\W\featmap(\coldict))^{-1}\featmap(i)
= \featmap(i)^\transp(\I/\alpha + \featmap(\coldict)^\transp\W\featmap(\coldict))^{-1}\featmap(i)\\
&\leq \tfrac{1}{1-\varepsilon} \featmap(i)^\transp(\I/\alpha + \featmap([n])^\transp\featmap([n]))^{-1}\featmap(i)
= \tfrac{1}{1-\varepsilon} \alpha\featmap(i)^\transp(\I + \alpha\featmap([n])^\transp\featmap([n]))^{-1}\featmap(i)
= \ell_i(\L).
\end{align*}

\textbf{Caching strategy.} Note that if we invoke \alphadpp multiple times for a fixed $\alpha$, we do not need to recompute all approximations $l_i$ from scratch each time. Rather,
we first store an eigendecomposition of $\Lbh$ to be able to quickly compute $(\alpha\L_{\coldict,\coldict} + \W^{-1})^{-1}$ in quadratic rather than cubic time.
Then, for each item $i$ we store a \emph{cache} of the current upper bound,
which is initialized to $\alpha\kappa^2$ and then lowered to $l_i$
when $l_i$ is actually computed. This way we never need to recompute
the same $l_i$ twice, and the runtime improves. In particular,
computing a single marginal $l_i$ requires $\cO(k^6)$ time.
So, if all $l_i$ were computed from scratch, then the inner loop of
\Cref{alg:alpha-dpp-sampler} would require $\alpha_{\max}\kappa^2k n \cdot k^6$
to compute $\alpha_{\max}\kappa^2k n$ marginals $l_i$, one for each item in $\unifset$. On the other hand, computing all $l_i$ for all items once and for all
would require $n \cdot k^6$ time, and then sampling would be near-constant time using an appropriate multinomial sampler (see \cite{derezinski2019exact}).
In our case, using the caching strategy we can get the best of both worlds $\acO(\min\{\alpha_{\max}\kappa^2k, 1\}\cdot n \cdot k^6)$ since
we never compute any $l_i$ more than~once.
 \end{document}